# Bayesian Logic Programs

## Kristian Kersting and Luc De Raedt

*Institute for Computer Science, Machine Learning Lab*
*University of Freiburg,*
*Georges-Koehler-Allee, Building 079,*
*D-79110 Freiburg/Brg., Germany*
E-mail: {kersting,deraedt}@informatik.uni-freiburg.de

Bayesian networks provide an elegant formalism for representing and reasoning about uncertainty using probability theory. They are a probabilistic extension of propositional logic and, hence, inherit some of the limitations of propositional logic, such as the difficulties to represent objects and relations. We introduce a generalization of Bayesian networks, called Bayesian logic programs, to overcome these limitations. In order to represent objects and relations it combines Bayesian networks with definite clause logic by establishing a one-to-one mapping between ground atoms and random variables. We show that Bayesian logic programs combine the advantages of both definite clause logic and Bayesian networks. This includes the separation of quantitative and qualitative aspects of the model. Furthermore, Bayesian logic programs generalize both Bayesian networks as well as logic programs. So, many ideas developed in both areas carry over.

## 1. Introduction

Bayesian networks [Pea91] are one of the most important, efficient and elegant frameworks for representing and reasoning with probabilistic models. A single Bayesian network specifies a joint probability density over a finite set of random variables and consists of two components: (1) a *qualitative* one that encodes the local influences among the random variables using a directed acyclic graph, and (2) a *quantitative* one that encodes the probability densities over these local influences. Bayesian networks have been applied to many real-world problems in diagnosis, forecasting, automated vision, sensor fusion and manufacturing control (cf. see the articles [HMW95,BH95,FF95,HBR95] which form together a special issue of the *Communications of the ACM*).

However, Bayesian networks are a probabilistic extension of propositional logic. The limitations of propositional logic, which Bayesian networks inherit, are well-known, see e.g. [Poo93,NH97,Jae97,FL98,Kol99]: they have a rigid structure and therefore have problems representing a variable number of objects or relations among objects. Consider e.g. building a probabilistic model of a class of computer networks with Bayesian networks. This is problematic because the complex and dynamic structure of computer networks, and the relations among their different components, cannot elegantly be modeled using Bayesian networks. Indeed, it is quite likely that the structure of different networks is at an abstract level quite similar. However, using Bayesian networks each computer network would need to be modeled by its own specific Bayesian network. There is no way of formulating general probabilistic regularities for all the computer networks. Furthermore, whenever components are added or deleted to a computer network its corresponding Bayesian network should be modified. This in turn would lead to exponential updating problems.

The above sketched problems are due to the propositional nature of Bayesian networks. They would disappear when using a first order formalism. It is therefore no surprise that various researchers have proposed first order extensions of Bayesian networks, e.g. [Poo93,NH97,Jae97,Kol99]. Many of these techniques employ the notion of *knowledge-based model construction* [BGW94,Had99], where first-order rules with associated uncertainty parameters are used to generate specific Bayesian networks for particular queries. This is especially useful in domains where the number of relevant random variables depends on the specific problem, such as computer networks, pedigree analysis, etc.

The main contribution of this paper is the introduction of Bayesian logic programs. Bayesian logic programs combine Bayesian networks with definite clause logic, i.e. "pure" Prolog. Therefore, Bayesian logic programs are easy to understand and use by practitioners in both communities. Indeed, both Bayesian networks and "pure" Prolog programs can naturally be represented using Bayesian logic programs. Bayesian logic programs can also handle domains involving structured terms as well as continuous random variables, which is not the case with the earlier proposals. Furthermore, whereas the approaches of [Poo93,NH97,Jae97,Kol99] view a ground atom as a state of a random variable Bayesian logic programs view them as random variables. More precisely, we establish a one-to-one mapping between ground atoms and random variables. This results − as for Bayesian networks − in a strict separation of quantitative and qualitative components of the representation, which is considered one of the key features of Bayesian networks.

We proceed as follows. Section 2 presents our example domain and reviews the basics of probability theory and definite clause logic. Section 3 motivates and introduces the representation language of Bayesian logic programs. Their declarative semantics are given in Section 4, and in Section 5 we discuss a query-answering procedure. Section 6 analysis the connection of the query-answering procedure to SLD trees. In Section 7 we investigate the representational power of Bayesian logic programs by showing how they can represent Bayesian networks, definite clause programs and dynamic Bayesian networks a.o. We discuss related work in Section 8 and conclude the paper in Section 9. The appendix contains the proofs of some theorems as well as simple Prolog shell for interpreting Bayesian logic programs.

## 2. Preliminaries

Bayesian logic programs combine Bayesian networks with definite clause logic. Throughout the paper, we assume some familiarity with logic programming or Prolog (see e.g. [SS86,Bra86,Fla94]) as well as with Bayesian networks (see e.g. [RN95,Nil98,Pea91,CDLS99]). We will now briefly review the key concepts and ideas underlying Bayesian networks and definite clause logic. Before doing so, we provide a motivating example.

### 2.1. A motivating example: genetics

Throughout the paper we will employ an example from the field of genetics to illustrate Bayesian logic programs. Genetics provide an intuitive and natural application domain for first order probabilistic models, because it

- has a probabilistic nature given by the biological laws of inheritance, and
- requires the representation of the relational familial structure of the individuals under study.

As for the computer networks example, the genetic and probabilistic regularities hold across different families though their structure may be different. Relational or first order representations can be used to capture the qualitative aspects. A second reason for using an illustration from the field of genetics for the purposes of this paper, is that genetics studies the inheritance of both continuous and discrete phenotypes such as weight and blood type. An individual's *phenotype* is an observable characteristic, whose distribution is affected by an individual's heritable information in conjunction with environmental factors.

The subfield of genetics which investigates continuous phenotypes is called *quantitative genetics*[1], cf. [WJSR80,Fal81,Tho86]. For a range of models within quantitative genetics "*each individual has a polygenic value, or polygenotype, which in the population is normally distributed*" [Tho86, Section 6.2]. These models are called *polygenic* models. The reason for that is that the effects of some genes on a phenotype seem to be additive. I.e. each gene independently effects additive changes of the phenotype. If equal effects are assumed, then the phenotype value is normally distributed as illustrated in Figure 1. Here, the genetic information $aa$ has no effect on the phenotype and $AA$ has an additive change of 2. Thus, the distribution of the phenotype is given by counting how often a particular additve effect might occur. We will use a simplified model for the polygentic value height as running example throughout the paper. Let $\mathcal{N}(x, \mu, \Sigma)$ denote a Gaussian density with mean $\mu$ and Variance $\Sigma$:

$$\mathcal{N}(x, \mu, \Sigma) := \frac{1}{\sqrt{2\pi\Sigma}} e^{-\frac{1}{2}\left(\frac{(x-\mu)^2}{\Sigma}\right)}.$$

---

[1] According to Falconer [Fal81, page 2], parts of the theoretical basis of quantitative genetics was established by the geneticist Sewal Wright, the same person to whom the idea of using graphical representations of probabilistic information can be traced back [Pea91].

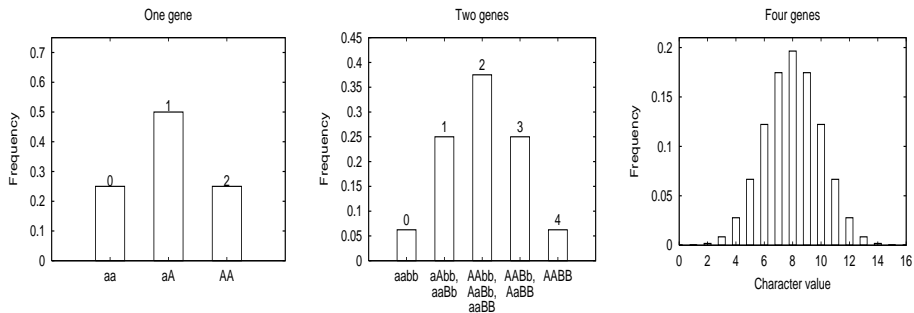

Figure 1. The effects of one, two and four different genes underlying a phenotype, e.g. height: when the number of underlying genes increases, the values of the phenotype tend to be normally distributed. The numbers associated to the boxes in the two left histograms are values of the phenotype. It is assumes that each gene independently but equally effects additive changes of the phenotype. Lower cases stand for *no effect*, and upper cases for *effect*.

Assume that the height of an individual has apriori a normal density with mean 175 [cm] and variance 60 [cm$^2$], i.e. $p(Height = h) = \mathcal{N}(h, 175, 60)$. Its aposteriori density depends on the heights of the individual's parents, $Height\_M$ and $Height\_F$:

$$p(Height = h \mid Height\_M = m, Height\_F = f) = \mathcal{N}(h, \frac{1}{2}(m + f), 60) \qquad (2.1)$$

The densities are visualized in Figure 2. Questions one could be interested in are e.g.: "what is the expected height of a person given that her grandfather's height is 182 cm ?" or "what is the expected height of a person given that her great granddaughter's height is 161 cm ?"

### 2.2. Bayesian networks

In the discussion of Bayesian networks we will use $X$ to denote a random variable, $x$ a state and $\mathbf{X}$ (resp. $\mathbf{x}$) a vector of variables (resp. values). Throughout the paper, the set $D(X)$ of all states of random variables $X$ is called the *domain* of $X$. The set $D(\mathbf{X}) := \bigotimes_{X \in \mathbf{X}} D(X)$, where $\bigotimes$ denotes the Cartesian product, is the domain, the set of *joint states* of the random variables $\mathbf{X}$. We will focus on *real* random variables of which *discrete* and *finite* random variables are special cases[2]. A (real) random variable can take values in the continuum, i.e. $D(X) = \mathbb{R}$, and we can only talk about the probability

$$P(X \in [a, b)) = \int_a^b p(X = x)\, dx \qquad (2.2)$$

that the state of $X$ falls in some interval $[a, b)$. The function $p : \mathbb{R} \mapsto \mathbb{R}$ is a *probability density*, i.e. $p(X = x) \geq 0$ for all $x \in D(X)$ and $\int_{-\infty}^{+\infty} p(X = x)\, dx = 1$. We will use the normal letter $p$, e.g. $p(X = x)$, to denote that $X$ takes the value $x \in D(X)$, and the bold letter $\mathbf{p}$ to denote a probability density, e.g. $\mathbf{p}(X)$. We will use the bold letter $\mathbf{P}$ to denote a *probability measure*, e.g. $\mathbf{P}(X)$. Given two real random variables $X, Y$ the conditional probability density of $X$ given $Y$ is defined as $\mathbf{p}(\mathbf{X} \mid \mathbf{Y}) = \frac{\mathbf{p}(\mathbf{X}, \mathbf{Y})}{\mathbf{p}(\mathbf{Y})}$. The *marginalized* probability density of $X$ given $\mathbf{p}(\mathbf{X}, \mathbf{Y})$ is $\mathbf{p}(\mathbf{X}) = \int_{-\infty}^{+\infty} \mathbf{p}(\mathbf{X}, \mathbf{Y} = \mathbf{y})\, d\mathbf{y}$. We denote the cardinality of a set $S$ with $|S|$.

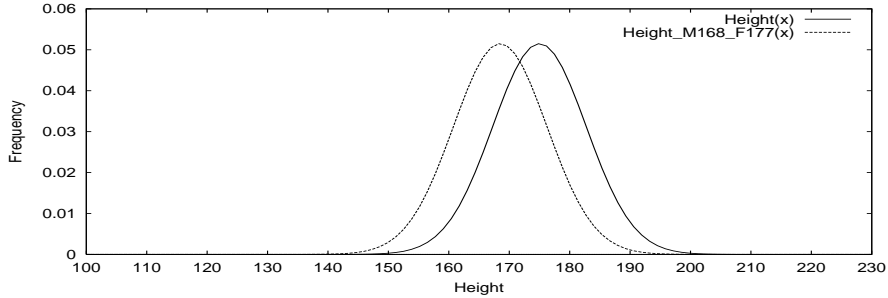

Figure 2. Example densities of the model of height inheritance. (1) The apriori density $p(Height = h)$ of a person's height is a normal density with mean 175 and variance 60, $p(Height = h) = \mathcal{N}(h, 175, 60)$. (2) The aposteriori density given that the heights $m, f$ of the parents $M, F$ are 164 and 173, $p(Height = h \mid Height\_M = m, Height_F = f) = \mathcal{N}(h, 168.5, 60)$.

A *Bayesian network* [Pea91,CDLS99] represents the joint probability density $\mathbf{p}(X_1, \ldots, X_n)$ (and due to equation (2.2) a probability measure) over a fixed, finite set $\{X_1, \ldots, X_n\}$ of random variables. It is an augmented, acyclic graph, where each node corresponds to a random variable $X_i$ (we will not distinguish between the random variables and the nodes of the graph) and each edge indicates a direct influence among the random variables. Figure 3 shows the graph of a Bayesian network modelling our height example for a particular family. The familial relationship, which is taken from the stud farm example in [Jen96], forms the basis for the graph. The network states e.g. that Irene's height is influenced by the heights of its parents Gwenn and Eric. The domain of each $X_i$, the set

---

[2] Generalizations to more 'complex' random variables such as $d$-dimensional real random variables could easily be obtained modulo the constraints well-known from probability theory (see e.g. [Bau91]).

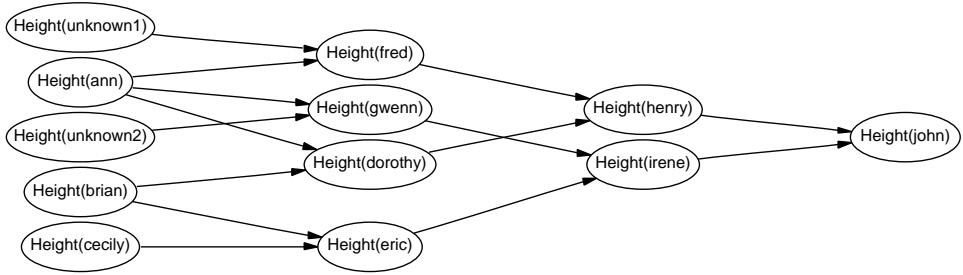

Figure 3. The graphical structure of a Bayesian network modelling the inheritance of height within a particular family. The familial relationship is taken from Jensen's stud farm example [Jen96, page 38].

of possible states of $X_i$ is $\mathrm{D}(X_i) = \mathbb{R}$. If we use a wildcard $Y$ for the individuals $ann, fred, \ldots$ this could be written $\mathrm{D}(Height(Y)) = \mathbb{R}$. The direct predecessors of a node $X$, the parents of $X$ are denoted by $\mathbf{Pa}(X)$. A Bayesian network stipulates a conditional independency assumption:

**Assumption 2.1** (independency). Each node $X_i$ in the graph is conditionally independent of any subset $\mathbf{A}$ of nodes that are not descendants of $X_i$ given a joint state of $\mathbf{Pa}(X_i)$, i.e. $\mathbf{p}(X_i \mid \mathbf{A}, \mathbf{Pa}(X_i)) = \mathbf{p}(X_i \mid \mathbf{Pa}(X_i))$.

E.g. $Height(irene)$ is conditionally independent of $Height(ann)$ given a joint state of its parents $\{Height(gwenn), Height(eric)\}$. Any pair $(X_i, \mathbf{Pa}(X_i))$ is called the *family* of $X_i$, e.g. $Height(irene)$'s family is $(Height(irene), \{Height(gwenn), Height(eric)\})$. Given the conditional independence assumption, we can write down the joint probability density as

$$\mathbf{p}(X_1, \ldots, X_n) = \prod_{i=1}^{n} \mathbf{p}(X_i \mid \mathbf{Pa}(X_i)) \qquad (2.3)$$

by applying the independency assumption 2.1 to the chain rule expression of the joint probability density. Thereby, we associate to each node $X_i$ of the graph the conditional probability densities $\mathbf{p}(X_i \mid \mathbf{Pa}(X_i))$, denoted as $cpd(X_i)$. The function $cpd(X_i)$ specifies for each $\mathbf{u} \in D(\mathbf{Pa}(X_i))$ and for each $u \in \mathrm{D}(X_i)$ the conditional density value $p(X_i = u \mid \mathbf{Pa}(X_i) = \mathbf{u})$, denoted as $cpd(X_i)(u \mid \mathbf{u})$. For our height example we set according to Equation (2.1):

$$cpd(X_i)(u) = \mathcal{N}(u, 175, 60), \qquad (2.4)$$

if $\mathbf{Pa}(X_i) = \{\}$, and

$$cpd(X_i)(u \mid u_f, u_m) = \mathcal{N}(u, \frac{1}{2}(u_m + u_f), 60) \qquad (2.5)$$

otherwise, where $u_m$ and $u_f$ are the heights of $X_i$'s mother and father. The *inference problem* for a Bayesian network could be stated as follows:

**Definition 2.2** *(inference problem). Given a Bayesian network over* $\{X_1, \ldots, X_n\}$, *a joint state* **u** *of* **Y** $\subset$ $\{X_1, \ldots, X_n\}$ *and a set of query variables* **Q** $\subset \{X_1, \ldots, X_n\}$, *find the conditional density* $\mathbf{p}(\mathbf{Q} \mid \mathbf{Y} = \mathbf{u})$.

In our height example, one might e.g. query for $p(Height(john))$ or $p(Height(john) \mid Height(ann) = 165)$. The answers would be $p(Heigth(john) = h) = \mathcal{N}(h, 175, 112.45)$ and $p(Height(john = h) \mid Height(ann) = 165) = \mathcal{N}(h, 171.25, 111.56)^3$.

The network in Figure 3 is an example of a *Gaussian network*, where all associated densities are Gaussian. In *discrete networks* the random variables are discrete, and *conditional Gaussian* (CG) networks (cf. [CDLS99]) are networks involving both discrete and continuous variables. In CG networks every discrete variable only depends on discrete variables, and continuous variables follow a multivariate Gaussian distribution given the discrete. In general, exact solutions of the inference problem are not possible when arbitrary conditional densities are employed. This is why CG networks are important. They constitute an analytically tractable model involving continuous and discrete variables. Many exact algorithms solving the inference problem for the tractable cases exist. Their details are not relevant for the present paper and we refer to literature (e.g. [Pea91,RN95,CDLS99]). Furthermore, approximate solutions of models involving arbitrary conditional densities can always be computed using stochastic techniques such as Gibbs sampling (see e.g. [JKK95]).

### 2.3. Definite Clause Logic

Imagine another family totally separated from the described one. A similar Bayesian network would model the height example within that family: its graphical structure and associated conditional probability distribution are controlled by the same intensional regularities. Definite clause logic is a classical framework for representing such (logical) intensional regularities.

A first-order alphabet is a set of predicate symbols and a set of functor symbols. Constants are functor symbols of arity 0. We assume that at least one constant is given. A *definite clause* is a formula of the form $A \leftarrow B_1, \ldots, B_m$ where $A$ and the $B_i$ are logical atoms. An *atom* $r(t_1, \ldots, t_n)$ is a predicate symbol $p$ followed by a bracketed $n$-tuple of terms $t_i$. A *term* $T$ is a variable $V$ or a functor symbol $f(t_1, \ldots, t_k)$ immediately followed by a bracked $n$-tuple of term $t_i$. A definite clause can be read as $A$ if $B_1$ and ... and $B_m$. All variables in definite clauses are universally quantified, although this is not explicitly written. We call $A$ the *head* and $B_1, \ldots, B_m$ the *body* of the clause. A *definite clause program* is a set of definite clauses. Consider e.g. the program *height* given in Figure 2.3. The motivation for this particular choice of program will become clear soon. Its alphabet consists of the predicate symbol *height* and the function symbols

---

[3] The answers are computed using the Hugin expert system [Huga].

$$height(ann) \leftarrow$$
$$height(cecily) \leftarrow$$
$$height(brian) \leftarrow$$
$$height(unknown1) \leftarrow$$
$$height(unknown2) \leftarrow$$
$$height(dorothy) \leftarrow height(ann), height(brian)$$
$$height(eric) \leftarrow height(cecily), height(brian)$$
$$height(fred) \leftarrow height(ann), height(unknown1)$$
$$height(gwenn) \leftarrow height(ann), height(unknown2)$$
$$height(henry) \leftarrow height(dorothy), height(fred)$$
$$height(irene) \leftarrow height(eric), height(gwenn)$$
$$height(john) \leftarrow height(henry), height(irene)$$

Figure 4. The definite clause program *height*.

*unknown1*, *unknown2*, *ann*, *brian*, *cecily*, *fred*, ... of arity 0. The first six clauses are called *facts*, because they have empty bodies.

The set of variables in a term, atom or clause $E$, is denoted as $vars(E)$. A clause is called *range-restricted*, if all variables occurring in the head of a clause also occur in its body, i.e. $vars(head(E)) \subseteq vars(head(E))$. All clauses of our example program are range-restricted. The following clause $C$ is not range-restricted

$$height(X) \leftarrow height(Y).$$

*Functor-free* clauses are clauses that contain only variables and constants as terms. A *goal* is a formula of the form $\leftarrow B_1, \ldots, B_m$. A substitution $\theta = \{V_1/t_1, \ldots, V_n/t_n\}$, e.g. $\{X/ann\}$, is an assignment of terms $t_i$ to variables $V_i$. Applying a substitution $\theta$ to a term, atom or clause $e$ yields the instantiated term, atom, or clause $e\theta$ where all occurrences of the variables $V_i$ are simultaneously replaced by the term $t_i$, e.g. $C\{X/ann\}$ yields $height(ann) \leftarrow height(Y)$ with $head(C)\{X/ann\} = height(ann)$. A substitution $\theta$ is called a *unifier* for a finite set $S$ of atoms if $S\theta$ is singleton. A unifier $\theta$ for $S$ is called a *most general unifier* (MGU) for $S$ if, for each unifier $\sigma$ of $S$, there exists a substitution $\gamma$ such that $\sigma = \theta\gamma$. The substitution $\{X/ann\}$ is the MGU of $\{height(ann), height(X)\}$. A term, atom or clause $E$ is called *ground* when there is no variable occurring in $E$, i.e. $vars(E) = \emptyset$. All clauses in our example program *height* are ground.

The *Herbrand base* of a definite clause program $T$, denoted as HB($T$), is the set of all ground atoms constructed with the predicate and functor symbols in the first-order alphabet of $T$. A *Herbrand interpretation* is a subset of HB($T$).

The *least Herbrand model* LH($T$) of a definite clause program is defined as the set of all facts $f \in$ HB($T$) such that $T$ logically entails $f$, i.e. $T \models f$. The least Herbrand model LH($T$) captures the semantics of the program $T$. For the above program *height*, we have that HB(*height*) = LH(*height*), which consists of all the ground atoms occurring in *height*.

## 3. Merging Bayesian Networks With Definite Clause Logic

Here, we show how Bayesian networks and definite clause logic are integrated within the framework of Bayesian logic programs. Before defining Bayesian logic programs formally, we present the key ideas and illustrate them on the height example.

### 3.1. *"Propositional" Bayesian Logic Programs*

Besides the papers [Poo93,NH97,Jae97,Kol99] the book of Pat Langley [Lan95] gives us a hint of how to combine Bayesian networks and definite clause logic. Langley does not represent Bayesian networks graphically but rather uses the notation of propositional definite clause programs. In the spirit of Langley, "propositional" Bayesian logic programs will represent the Bayesian network of Figure 3 using the clauses of the definite clause program *height* in Figure 2.3: each family $(X, \mathbf{Pa}(X))$ of the Bayesian network corresponds to a clause $X \leftarrow \mathbf{Pa}(X)$ in Langley's notation. As a consequence, each random variable of the Bayesian network corresponds to a predicate (with arity 0, i.e. a proposition) in the definite clause program. Furthermore, for each random variable there will be exactly one clause whose head contains the random variable. On our example, one can now easily verify that the graphical structure of the Bayesian network corresponds to the *dependency graph* of the definite clause program. The dependency graph $DG(T)$ of a definite clause program $T$ is the directed graph whose nodes correspond to the ground atoms in LH($T$)[4]. The graph has an edge from a node $A$ to a node $B$ if and only if there exists a clause $C \in T$ and a substitution $\theta$, such that $B = head(C)\theta$, $A \in body(C)\theta$ and for all atoms $A' \in C\theta$ : $A' \in LH(T)$. The dependency graph $D(height)$ for our example program *height* is given in Figure 3 which shows the Bayesian network of the height example. Thus, the dependency graph captures the *qualitative* component of a Bayesian network. To capture the *quantitative* component, we have to associate to each clause $X \leftarrow \mathbf{Pa}(X)$ the conditional probability densities $cpd(X)$ of the corresponding random variable $X$. Similarly, we will carry over the domains $D(X)$ of the random variables $X$ to the predicates of the Bayesian logic program. This actually implies that the predicates and clauses in the Bayesian logic program can be interpreted in two different manners. First, there is the *logical* interpretation, which interpretes the clauses and predicates as (propositional) definite clauses and logical predicates. Secondly, there is the *Bayesian* interpretation, which considers the (propositional) definite clause program as a Bayesian network and the (ground) atoms as the random variables.

---

[4] The usual definition (see e.g. [AHV95]) considers the ground atoms in the Herbrand base. We use the least Herbrand model in order to put the idea of Bayesian logic programs across.

The key idea underlying Bayesian logic program is now that we generalize Langley's notation towards first order logic by considering first order definite clauses instead of propositional ones, i.e. to allow for logical variables and to interpret the dependency graph of a definite clause program as the graphical structure of a Bayesian network.

### 3.2. Bayesian Logic Programs

Let us now define the key concept of Bayesian logic programs.

**Definition 3.1** *(Bayesian (definite) clause). A Bayesian (definite) clause c is an expression of the form*

$$A \mid A_1, \ldots, A_n \qquad (3.1)$$

*where $n \geq 0$, the $A, A_1, \ldots, A_n$ are Bayesian atoms and all atoms are (implicitly) universally quantified.*

So, the differences between a *Bayesian* and a *definite* clause are :

- the atoms $r(t_1, ..., t_n)$ and predicates are *Bayesian*, which means that they have an associated domain $D(r)$, and
- we use " $\mid$ " instead of ":-" to capture the idea of conditional probability densities.

Note that the domain $D(r)$ is unique for the predicate $r$. Furthermore, all other logical notions carry over to Bayesian logic programs. So we will speak of Bayesian predicates, terms, constants, functors, substitutions, ground Bayesian clauses, etc. Throughout the paper, we will use Prolog notation to write down clauses. Variables start with a capital; constant, functor and predicate symbols start with a lowercase.

**Example 3.2.** The Bayesian clause $c$

$$\texttt{height(X)} \mid \texttt{mother(Y,X), father(Z,X), height(Y), height(Z)} \qquad (3.2)$$

defines the height of an individual `X` in terms of the heights of its mother `Y` and its father `Z`. The Bayesian predicates of $c$ are `height, mother` and `father`. The domains of these predicates are $D(father) = D(mother) = \{true, false\}$ and $D(height) = \mathbb{R}$. $\qquad \square$

Intuitively, a Bayesian predicate generically represents a set of random variables, and each Bayesian ground atom uniquely represents a random variable. To each Bayesian clause $c$ we associate the *conditional probability densities $cpd(c)$*.

| $\mathtt{mother}(Y,X)=u_1$ | $\mathtt{father}(Z,X)=u_2$ | $cpd(c)(h \mid u_1,u_2,u_3,u_4)$ |
|:---:|:---:|:---:|
| *true* | *true* | $\mathcal{N}(h,\frac{1}{2}(u_3+u_4),60)$ |
| *true* | *false* | $\mathcal{N}(h,u_3,60)$ |
| *false* | *true* | $\mathcal{N}(h,u_4,60)$ |
| *false* | *false* | $\mathcal{N}(h,175,60)$ |

Table 1

The conditional probability densities $cpd(c)$ associated to the Bayesian clause $c$ of example 3.2: $\mathtt{height(X)} \mid \mathtt{mother(Y,X)},\mathtt{father(Z,X)},\mathtt{height(Y)},\mathtt{height(Z)}$. The parameters $u_3, u_4$ refer to the heights of of individual $X$'s parents.

**Definition 3.3** (*associated conditional probability densities*). *Let $c$ be the Bayesian clause*

$$\mathtt{r(t_1,\dots,t_n)} \mid \mathtt{s_1(t_{1,1},\dots,t_{1,n_1})},\dots,\mathtt{s_m(t_{m,1},\dots,t_{m,n_m})}.$$

*The conditional probability densities $cpd(c)$ specify for each $(u_1,\dots,u_m) \in D(s_1) \times \dots \times D(s_m)$ a function $cpd(c)(u \mid u_1,\dots,u_m) : D(r) \mapsto [0,1]$ with*

$$cpd(c)(u \mid u_1,\dots,u_m) =$$
$$p(r(t_1,...,t_n) = u \mid s_1(t_{1,1},\dots,t_{1,n_1}) = u_1,\dots,s_m(t_{m,1},\dots,t_{m,n_m}) = u_m).$$

Because all Bayesian ground atoms $r\theta$ over a Bayesian predicate $r$ inherit their domains from $r$, i.e. $D(r\theta) := D(r)$, the densities $cpd(c)$ generically represent the conditional probability densities of all ground instances $c\theta$. Thus, $cpd(c)$ specifies the quantitative component of $c$.

**Example 3.4** (continuing example 3.2). The conditional probability densities $cpd(c)$ associated to the Bayesian clause $c$ of the last example are given in Table 1. Let $\theta = \{X \mapsto john, Y \mapsto irene, Z \mapsto henry\}$ be a substitution. The ground instance $c\theta$ specifies the conditional probability densities:

$$\mathbf{p}(height(john) \mid mother(irene, john), height(irene),$$
$$father(henry, john), height(henry)).$$

$\square$

At this point the reader may see some further connections to Bayesian networks. Indeed, reconsider the last example. The random variables $mother(irene, john), father(henry, john), height(irene), height(henry)$ directly influence $height(john)$. More generally any clause $c$ specifies a direct probabilistic influence of each ground atom in $body(c)\theta$ on $head(c)\theta$ for any ground instance $c\theta$, when all ground atoms in $c\theta$ are true.

So far, we have however ignored one important complication. When representing a Bayesian network as a set of propositional clauses, there will be exactly one clause that defines each Bayesian predicate (i.e. the clause containing the predicate in the head). In Bayesian logic programs, one may have two clauses $c_1$ and $c_2$ and corresponding substitutions $\theta_i$ that ground the clauses $c_i$ such that $head(c_1\theta_1) = head(c_2\theta_2)$. This can lead to problems as illustrated in the following example.

**Example 3.5.** Consider the Bayesian clauses

$$\texttt{height(X)} \,|\, \texttt{mother(Y,X)}, \texttt{height(Y)}.$$

$$\text{and } \texttt{height(X)} \,|\, \texttt{father(Y,X)}, \texttt{height(Y)}.$$

in the light of the substitutions $\theta_1 = \{X \leftarrow jef, Y \leftarrow mary\}$ and $\theta_2 = \{X \leftarrow jef, Y \leftarrow john\}$. The ground clauses $c_1\theta_1$ and $c_2\theta_2$ specify

$$\mathbf{p}(height(jef) \,|\, mother(mary, jef, ), height(mary))$$

$$\text{and } \mathbf{p}(height(jef) \,|\, father(john, jef, ), height(john))$$

but they do not specify the needed conditional probability densities:

$$\mathbf{p}(height(jef) \,|\, mother(mary, jef), height(mary),$$
$$father(john, jef), height(john))$$

$\square$

Notice at this point that the two clauses $c_1, c_2$ may be identical! The standard solution to obtain the distribution required are so called *combining rules*[5]. Our notion follows closely [NH97] and differs mainly in the restriction that the input set is finite. We make this assumption for computational reasons and will only weaken this restriction when embedding pure Prolog programs (cf. section 7.2).

**Definition 3.6** (*Combining rule*). *A combining rule is any algorithm that maps every finite set of conditional probability densities*

$$\{\mathbf{p}(A \,|\, A_{i1}, \ldots, A_{in_i}) \,|\, 1 \leq i \leq m, \ n_i \geq 0\}, \tag{3.3}$$

$m \geq 1$, *onto the conditional probability densities*

$$\mathbf{p}(A \,|\, B_1, \ldots, B_n), \tag{3.4}$$

*called the combined conditional probability densities, with* $\{B_1, \ldots, B_n\} = \bigcup_{i=1}^{m}\{A_{i1}, \ldots, A_{in_i}\}$ *and* $n < \infty$. *Its outputs are empty if and only if its inputs are empty.*

We have claimed the equality $\{B_1, \ldots, B_n\} = \bigcup_{i=1}^{m}\{A_{i1}, \ldots, A_{in_i}\}$ for the sake of simplicity[6]. It will make the specification of probabilistic influences among ran-dom variables, definition 4.4, and the formulation of an independency assumption, assumption 4.6, straight forward. In any case, combining rules can be seen as a generalization of the idea of canonical distributions (cf. [RN95, page 443]), that is, the relationship between the parents and a child fits some standard pattern, and of local probabilistic models, i.e. finer-grained structure within the associated conditional probability densities, see e.g. [BFGK96].

---

[5] Similar concepts are used in other proposals, e.g. combinations functions [Jae97] or aggregate functions [Kol99].

[6] This is no restriction. Assume $\mathbf{B} = \{B_1, \ldots, B_n\}$ to be a proper subset of $\mathbf{A} = \bigcup_{i=1}^{m}\{A_{i1}, \ldots, A_{in_i}\}$. Then, we replace the original combining rule with a combining rule that outputs the combined densities $\mathbf{f}(A \,|\, \mathbf{A})$ where $\forall \mathbf{c} \in \mathrm{D}(\mathbf{A} \setminus \mathbf{B}) : \mathbf{f}(A \,|\, \mathbf{B}, \mathbf{A} \setminus \mathbf{B} = \mathbf{c}) = \mathbf{p}(A \,|\, \mathbf{B})$. This is always possible.

**Example 3.7.** The functional formulation of the combining rule max, a rule which will be useful for embedding pure Prolog program into Bayesian logic programs, is

$$\max\{\mathbf{p}(A \mid A_{i1}, \ldots, A_{in_i}) \mid i = 1, \ldots, n\} = $$
$$\mathbf{p}(A \mid \cup_{i=1}^n \{A_{i1}, \ldots, A_{in_i}\}) := \max_{i=1}^n \{\mathbf{p}(A \mid A_{i1}, \ldots, A_{in_i})\}. \tag{3.5}$$

Another combining rule for Bayesian predicates having a boolean domain, e.g. $\{true, false\}$, is noisy_or. It is widely used in the *Uncertainty in AI* community. The rule noisy_or generalizes (see [RN95]) the logical *or* under three assumptions: (1) each cause has an independent chance of causing the effect, (2) all possible causes are listed and (3) whatever inhibits a parent $pa_1$ from causing the child is independent of whatever inhibits another parent $pa_2$, $pa_1 \neq pa_2$, from causing the child. Formally, the input $\{\mathbf{p}(A \mid A_i) \mid 1 \leq i \leq m\}$ over boolean random variables $A, A_1, \ldots, A_m$ is mapped onto $\mathbf{p}(A \mid A_1, \ldots, A_m)$ with

$$p(A = false \mid A_1 = a_1, \ldots, A_m = a_m) = \prod_{k=1}^m \langle 1 - p(A = a \mid A_k = true) \rangle^{a_i}$$
$$p(A = true \mid A_1 = a_1, \ldots, A_m = a_m) = 1 - p(A = false \mid A_1 = a_1, \ldots, A_m = a_m)$$

for $a_i \in D(A_i)$, $i = 1, \ldots, m$, and where

$$\langle x \rangle^a = \begin{cases} x & : & a = true, \\ 1 & : & a = false. \end{cases}$$

As a last illustration, we consider a real random variable $X$ in a CG network. A rule, which computes for each joint state of the discrete parents of $X$ a weighted sum of the states of the continuous parents of $X$ and sets this as mean of the Gaussian density of $X$, models a kind of regression. □

By now we are able to formally define the notion of a Bayesian logic program.

**Definition 3.8** *(Bayesian logic program). A Bayesian logic programs $B$ consists of a finite set of Bayesian clauses. To each Bayesian clause $c$ there is exactly one cpd($c$) associated, and for each Bayesian predicate $r$ there is exactly one associated combining rule comb($r$).*

**Definition 3.9** *(corresponding logic program). Let B be a Bayesian logic program. The set of logical definite clauses corresponding to the set of Bayesian clauses of B is called the corresponding logic program B.*

Attention should be paid to the fact that the definition allows for *functor symbols* to be used. Herein Bayesian logic programs differ from probabilistic relational models [Kol99] and relational Bayesian networks [Jae97] which restrict themselves to functor-free languages. This corresponds to using pure Prolog instead of datalog.

**Example 3.10.** The following Bayesian logic program *height* models our genetic domain (Section 4 will prove this):

```
father(unknown1,fred).  mother(ann, fred).  father(brian,dorothy).
mother(ann, dorothy).   father(brian,eric). mother(cecily,eric).
father(unknown2,gwenn). mother(ann,gwenn).  father(fred,henry).
mother(dorothy,henry).  father(eric,irene). mother(gwenn,irene).
father(henry,john).     mother(irene,john).

height(ann).            height(brian).  height(cecily).
height(unknown1).       height(unknown2).

height(X) | mother(Y,X), father(Z,X), height(Y), height(Z).
```

We associate to each Bayesian predicate the identity as combining rule and to each Bayesian ground fact over `mother` or `father` conditional probability densities of the form:

| $\mathbf{P}(mother(X,Y))$ | | | $\mathbf{P}(father(X,Y))$ | |
|---|---|---|---|---|
| *true* | *false* | and | *true* | *false* |
| 1.0 | 0.0 | | 1.0 | 0.0 |

The associated densities of each Bayesian ground fact over `height` resembles equations 2.4 and 2.5, and the associated densities of the remaining clause are given in example 3.4. □

To summarize, we have introduced Bayesian logic programs. They combine definite clause logic with Bayesian networks by establishing a one-to-one mapping between ground atoms and random variables. Thus, a logical variable remains a logical variable. This separates Bayesian logic programs from existing *knowledge-based model construction* approaches such as probabilistic logic programs [NH97], relational Bayesian networks [Jae97] and probabilistic relational models [Kol99]. In the latter frameworks, ground atoms represent states of random variables. Bayesian clauses generically specify possible direct influences. The associated conditional probability densities together with the associated combining rules probabilistically quantify these influences. Thus, Bayesian logic programs nicely separate the quantitative and the qualitative components.

## 4. Declarative Semantics

Intuitively, each Bayesian logic program $B$ specifies a (possibly infinite) Bayesian network, i.e. a joint probability density over a countable set of random variables. This view implicitly assumes that all knowledge about the domain of discourse is encoded in the Bayesian logic program (e.g. the horses belonging to a farm or to a pedigree). If the domain of discourse changes (e.g. the horses under consideration), then part of the Bayesian logic program has to be changed. Usually, these modifications will only concern Bayesian ground atoms (e.g. the Bayesian ground atoms over "mother", "father"). This is akin to the extensional facts of a database. The clauses then correspond to intensional rules.

The semantics of definite clause programs has been well-studied (see e.g. [Llo89]). The main result is that the intended meaning of a definite clause program, such as $\widetilde{B}$, is represented by its least Herbrand model $\mathrm{LH}(\widetilde{B}) \subset \mathrm{HB}(\widetilde{B})$ which contains all ground atoms of the Herbrand base $\mathrm{HB}(\widetilde{B})$ that are logically entailed by the program. If we ignore termination issues, they can - in principle - be computed by a theorem prover, such as e.g. Prolog. Various methods exist to compute the least Herbrand model. We merely sketch its computation through the use of the well-known *immediate consequence* operator $T_{\widetilde{B}}$ (cf. [Llo89]). For simplicity, we will assume that all clauses in a Bayesian logic program are range-restricted. A clause is *range-restricted* iff all variables occurring in the head also occur in the body. Range restriction is often imposed in computational logic. It allows to avoid derivation of non-ground true facts, i.e all facts entailed by the program are ground.

**Definition 4.1** *(immediate consequence operator). Let $B$ be a Bayesian logic program and $\mathcal{I}$ a Herbrand interpretation over $\widetilde{B}$. The immediate consequence operator $T_{\widetilde{B}}$ defined by $\widetilde{B}$ is the function on the set of all Herbrand interpretations of $\widetilde{B}$ such that for any such interpretation $\mathcal{I}$ we have*

$$T_{\widetilde{B}}(\mathcal{I}) = \{ A\theta \mid there\ is\ a\ substitution\ \theta\ and\ a$$
$$clause\ A \mid A_1, \ldots, A_n\ in\ \widetilde{B}\ such\ that$$
$$A\theta \mid A_1\theta, \ldots, A_n\theta\ is\ ground\ and$$
$$for\ all\ i \in \{1, \ldots, n\}\colon A_i\theta \in \mathcal{I}\}.$$

**Definition 4.2** *(least Herbrand model). Let $B$ be a Bayesian logic program. The least Herbrand model $\mathrm{LH}(\widetilde{B})$ is defined as the least fixpoint of $T_{\widetilde{B}}$ applied on $\emptyset$, i.e. $T_{\widetilde{B}}(\mathrm{LH}(\widetilde{B})) = \mathrm{LH}(\widetilde{B}) = T_{\widetilde{B}}(T_{\widetilde{B}}(\ldots T_{\widetilde{B}}(\emptyset) \ldots)$.*

**Example 4.3.** The least Herbrand model of the corresponding logic program $\widetilde{B}$ of the Bayesian logic program in example 3.10 coincides with $\mathcal{I}_2$ because

$$
\begin{aligned}
\mathcal{I}_1 := T_{\widetilde{B}}(\emptyset) &= \text{the set of all ground facts in } \widetilde{B} \\
\mathcal{I}_2 := T_{\widetilde{B}}(\mathcal{I}_1) &= \mathcal{I}_1 \cup \{\, height(fred),\, height(dorothy), \\
& \qquad\quad height(eric),\, height(gwenn), \\
& \qquad\quad height(herny),\, height(irene), \\
& \qquad\quad height(john)\} \\
\mathcal{I}_3 := T_{\widetilde{B}}(\mathcal{I}_2) &= \mathcal{I}_2 \\
& \;\;\vdots
\end{aligned}
$$

$\square$

Now, due to the one-to-one mapping between logical ground atoms, Bayesian ground atoms and random variables there exists exactly one set of random variables corresponding to $\mathrm{HB}(\widetilde{B})$ as well as exactly one set corresponding to $\mathrm{LH}(\widetilde{B})$. We define the former set as the Herbrand base $\mathrm{HB}(B)$ and the latter set as the least Herbrand model $\mathrm{LH}(B)$ of the Bayesian logic program $B$. As for definite clause programs $\mathrm{HB}(B)$ constitutes all random variables we can talk about given $B$, and $\mathrm{LH}(B)$ specifies the *proper* random variables; these are the ones for which (conditional) probability densities are (well-) defined. Much that, all other *logical* notions carry over to Bayesian logic programs. So, we will speak of Bayesian predicates, terms, constants, substitutions, ground Bayesian clauses, dependency graph etc.

So far, we have only characterized the proper random variables of discourse, i.e. the nodes in the Bayesian network. What is left is to introduce local influences among them, i.e. the edges in the Bayesian network. A natural candidate as medium for this is the already used *immediate consequence* operator.

**Definition 4.4** (*direct influence*[7]). *Let $B$ be a Bayesian logic program. A random variable $C$ directly influences a random variable $A$ if and only if*

1. $A, C \in \mathrm{LH}(B)$ *and*

2. *there is a Bayesian clause $A' \mid A_1, \ldots, A_n$ in $B$ and a substitution $\theta$ that grounds the clause such that $A = A'\theta$ and $C = A_i\theta$ for some $i$, $1 \le i \le n$, and $A_j\theta \in \mathrm{LH}(B)$ for all $1 \le j \le n$.*

*The set of random variables directly influencing a random variable $A \in \mathrm{LH}(B)$ is denoted as $\mathbf{Pa}(A)$, the parents of $A$. The recursive closure of the direct influence relation over $\mathrm{LH}(B)$ defines the influence relation.*

Roughly speaking, a random variable $C$ influences a random variable $A$ whenever there is a proof of $A$ that relies on $C$.

**Example 4.5** (continuing our running example). Figure 5 shows the *direct influence* relation of the height example. The filled nodes correspond to the nodes of the Bayesian network in Figure 3. $\square$

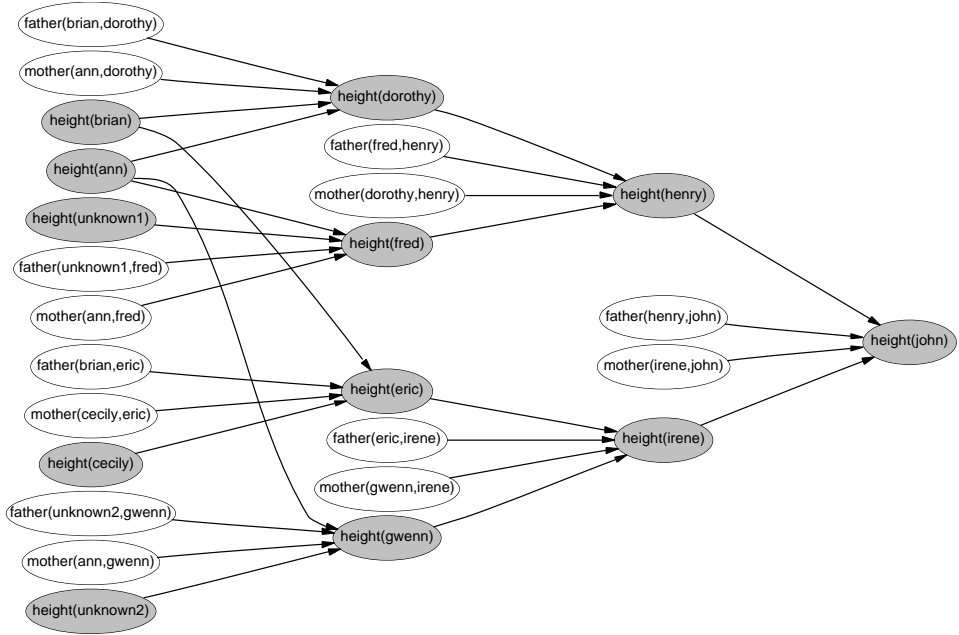

Figure 5. The *direct influence* relation induced over the least Herbrand model of the Bayesian logic program *height* of example 3.10. The filled nodes correspond to the nodes of the Bayesian network of Figure 3 modelling the stud farm example.

An induction over the cardinality of LH($B$) shows that the dependency graph $DG(B)$ graphically represents the *influence by* relation. Using the *influenced by* relation we are able to state a conditional independency assumption similar to that of Bayesian networks, cp. assumption 2.1:

**Assumption 4.6** (independency[7]). Each node $A \in \mathrm{LH}(B)$ in the dependency graph $DG(B)$ is conditionally independent of any subset $\mathbf{A} \subset \mathrm{LH}(B)$ of nodes that are not descendants of $A$ given a joint state of $\mathbf{Pa}(A)$, i.e. $\mathbf{p}(A \mid \mathbf{A}, \mathbf{Pa}(A)) = \mathbf{p}(A \mid \mathbf{Pa}(A))$.

**Example 4.7.** The *direct influence* relation of our running example shown in Figure 5 encodes e.g. the independency

$$\mathbf{p}(height(john) \mid mother(irene, john), height(irene),$$
$$father(henry, john), height(henry), \mathbf{C})$$
$$= \mathbf{p}(height(john) \mid mother(irene, john), height(irene),$$
$$father(henry, john), height(henry)),$$

---

[7] Without the claimed equality in the definition 3.6 of combining rules one would have to take the *combined* conditional probability densities into account. The *influenced by* relation would become a subset of currently employed one.

where **C** is any set of variables influencing *height(john)*. □

So far, we have ignored one important requirement: the *influenced by* relation should be acyclic in order to obtain a well-defined Bayesian network. I.e. the dependency graph $DG(B)$ should be acyclic (in the usual graph theoretical sense). Now, the network or graph can only be cyclic if there exists a proper random variable that influences itself. However, if there exists such a proper random variable $A$, executing the query ?- $A$. (using Prolog) would also be problematic. The SLD tree (cf. see [Llo89] and Section 6) of the query would be infinite and the query may not terminate[8]. Thus in such cases the logical component of the Bayesian logic program is itself problematic. With this in mind, we can formulate Theorem 4.9. The idea is to interpret the dependency graph of a Bayesian logic program $B$ as the graphical structure of a (possibly infinite) Bayesian network.

**Definition 4.8** *(well-defined Bayesian logic program). Let $B$ be a Bayesian logic program. If*

1. $\mathrm{LH}(B) \neq \{\}$,

2. *the dependency graph $DG(B)$ is acyclic, and*

3. *each random variable in $\mathrm{LH}(B)$ is only influenced by a finite set of random variables,*

*then $B$ is called* well-defined.

The *height* Bayesian logic program in example 3.10 is well-defined.

**Theorem 4.9** *(declarative semantics). Every well-defined Bayesian logic program $B$ specifies a unique probability measure over $\mathrm{LH}(B)$.*

*Proof.* This proof can be skipped without loss of continuity. Background material and definitions from probability theory are introduced in Section A in the appendix.

Let $B$ be a well-defined Bayesian logic program. We can assume that all proper random variables are real random variables, i.e. $\forall A \in \mathrm{LH}(B) : \mathrm{D}(A) = \mathbb{R}$. The existence and the uniqueness of $\mathrm{LH}(B)$ is guaranteed [Llo89], and $\mathrm{LH}(B)$ is countable [Doe94]. We will show, that $B$ specifies a *projective family of probability measures* (cf. Appendix A): A family $(\mathbb{R}^I, \mathcal{R}^I, P_I)_{I \in \mathcal{H}(T)}$ (where $T$ is a non-empty, countable set) of probability spaces is called *projective* if for all $H, J \in \mathcal{H}(T)$ with $H \subset J$ the equation $P_J = proj_J^H(P_H)$ holds, where $\mathcal{H}(T)$ is the set of all non-empty, finite subsets of $T$, and $proj_J^H$ is the projection of $\mathbb{R}^H$ onto $\mathbb{R}^J$. Then, the existence of a unique probability measure over $\mathrm{LH}(B)$ follows from *Kolmogorov*'s theorem.

Because $B$ is well-defined, we can assign to each random variable $A \in \mathrm{LH}(B)$ a finite rank:

---

[8] Termination depends on the search strategy of the theorem prover.

1. $\mathrm{rank}(A) = 0$, if no random variable in $\mathrm{LH}(B)$ is influencing $A$.

2. $\mathrm{rank}(A) = \max\{\mathrm{rank}(D) \mid D \in \mathrm{LH}(B) \wedge D$ is directly influencing $A\} + 1$, otherwise.

Indeed, rank induces a total order $\pi$ over $\mathrm{LH}(B)$. So, let $(A_n)_{n \in T}$ (for some index set $T$) be the sequence of random variables in $\mathrm{LH}(B)$ in ascending order according to $\pi$. We show now that $B$ specifies for each $J \in \mathcal{H}(T)$ a unique probability space.

The set of random variables constituting the space desired are $\mathbf{A}(\mathbf{J}) = \{A_{j_1}, \ldots, A_{j_m}\}$, and, hence, the measurable space is $(\mathbb{R}^J, \mathcal{B}^J)$. Furthermore, $B$ defines for each $A \in \mathrm{LH}(B)$ exactly one associated (combined) conditional probability density $cpd(A \mid \mathbf{Pa}(A))$. These densities could be seen as constraints on the probability measure $P_J$ over $(\mathbb{R}^J, \mathcal{B}^J)$. For that purpose, take the completion $\mathbf{C}(\mathbf{J})$ of $\mathbf{A}(\mathbf{J})$ with respect to the *influenced by* relation over $\mathrm{LH}(B)$ into account:

$$\mathbf{C}(\mathbf{J}) := \{D \in \mathrm{LH}(B) \mid D \text{ influences some } A \in \mathbf{A}(\mathbf{J})\} \qquad (4.1)$$

The set $\mathbf{C}(\mathbf{J})$ is always finite because $B$ is well-defined. An induction over $|\mathbf{C}(\mathbf{J})|$ shows that $\mathbf{C}(\mathbf{J})$ together with the densities $cpd(D \mid \mathbf{Pa}(D))$, $D \in \mathbf{C}(\mathbf{J})$ uniquely induce a Bayesian network $N(J)$: the nodes of $N(J)$ are $\mathbf{C}(\mathbf{J})$ and its edges are given by the *influence* relation. The network $N(J)$ specifies the unique probability densities $p_{\mathbf{C}(\mathbf{J})}$ (and therefore a unique probability measure $P_{\mathbf{C}(\mathbf{J})}$) over $(\mathbb{R}^K, \mathcal{B}^K)$, $K \in \mathcal{H}(T)$, as follows:

$$p_{\mathbf{C}(\mathbf{J})}(\mathbf{C}(\mathbf{J})) = \prod_{A \in \mathbf{A}(\mathbf{J})} cpd(A \mid \mathbf{Pa}(A)). \qquad (4.2)$$

Now, the probability densities $p_J$ are marginalized densities of $p_{C(J)}$, i.e.

$$p_J(\mathbf{A}(\mathbf{J})) = \int_{-\infty}^{+\infty} \ldots \int_{-\infty}^{+\infty} p_{\mathbf{C}(\mathbf{J})}(\mathbf{C}(\mathbf{J})) \, d\mathbf{D} \qquad (4.3)$$

where $\mathbf{D} = \mathbf{C}(\mathbf{J}) \setminus \mathbf{A}(\mathbf{J})$. Let $P_J$ the probability measure uniquely specified by $p_J$. Then $(\mathbb{R}^J, \mathcal{B}^J, P_J)$ is the unique probability space desired.

The densities $p_{\mathbf{C}(\mathbf{J})}$ are the key to prove that the family $(\mathbb{R}^I, \mathcal{B}^I, P_I)_{I \in \mathcal{H}(T)}$ of probability spaces is projective. Let $H = \{h_1, \ldots, h_n\} \in \mathcal{H}(T)$ with $J \subset H$ and $\mathbf{A}(\mathbf{H}) = (A_{h_1}, \ldots, A_{h_n})$. Performing the same steps for $\mathbf{A}(\mathbf{H})$ as for $\mathbf{A}(\mathbf{J})$ we obtain a Bayesian network $N(H)$. An induction over $|N(J)|$ proves, that $N(J)$ is a subnetwork of $N(H)$. Thus

$$p_{\mathbf{C}(\mathbf{J})}(\mathbf{C}(\mathbf{J})) = \int_{-\infty}^{+\infty} \ldots \int_{-\infty}^{+\infty} p_{\mathbf{C}(\mathbf{H})}(\mathbf{C}(\mathbf{H})) \, d\mathbf{D}' \qquad (4.4)$$

where $\mathbf{D}' = \mathbf{C}(\mathbf{H}) \setminus \mathbf{C}(\mathbf{J})$. Because (4.2) and (4.3) applies to both $\mathbf{A}(\mathbf{J})$ and $\mathbf{A}(\mathbf{H})$ it follows that $p_J$ are marginalized densities of $p_H$, the densities over $(\mathbb{R}^H, \mathcal{B}^H)$. This together with the fact, that $N(J)$ is a subnetwork of $N(H)$, means that the family $(\mathbb{R}^I, \mathcal{B}^I, P_I)_{I \in \mathcal{H}(T)}$ of probability spaces is projective. It then follows from *Kolmogorov*'s theorem that there exists a unique probability measure $P_T$ over $(\mathbb{R}^T, \mathcal{B}^T)$ which satisfies all constraints given by the associated (combined) conditional probability densities. This proves the theorem because the total order $\pi$ could be any rank respecting total order. □

Not every Bayesian logic program is well-defined. Let us investigate some examples of *ill-defined* programs.

**Example 4.10.** The Bayesian logic program

```
r(X) | s(X).
```

is not well-defined because its least Herbrand model is empty. The program

```
r(a).
s(a,b).
r(X) | r(X).
r(X) | s(X,f(Y)).
s(X,f(Y)) | s(X,Y).
```

is ill-defined as well because the random variable $r(a)$ is directly influenced by $s(a, f(b)), s(a, f(f((b)))), \ldots$ and by itself. The Bayesian logic program

```
r(a).
s(a).
r(X)    | r(f(X)).
r(f(X)) | s(f(X)).
s(f(X)) | s(X).
```

is ill-defined, because the random variable $r(a)$ is influenced by $r(f(a)), r(f(f((a)))), \ldots$ though $r(a)$ has a finite proof. The *direct influence by* relations of the two latter programs are shown in Figure 6. □

For the rest of the paper, we will consider a Bayesian logic programs to be well-defined if nothing contrary is stated explicitly.

To summarize, as a consequence of the one-to-one mapping of ground atoms onto random variables, we can switch between logical concepts like ground atoms, *direct consequence* operator and concepts of probability theory like random variable and *direct influence*. The Herbrand base of a Bayesian logic programs constitutes the set of random variables and its least Herbrand model specifies the proper random variables of discourse. The *direct consequence* operator (and the combining rules) induces the direct influence relation over the proper variables. Thus, the dependency graph augmented with the combined conditional probability densities can in principle be interpreted as a (possibly infinite) Bayesian network. This holds under the reasonable assumption of well-definition, i.e. no random variable influences itself or is influenced by infinite many other random variable. We will sometimes refer to the augmented dependency graph as the Bayesian network of the given Bayesian logic program.

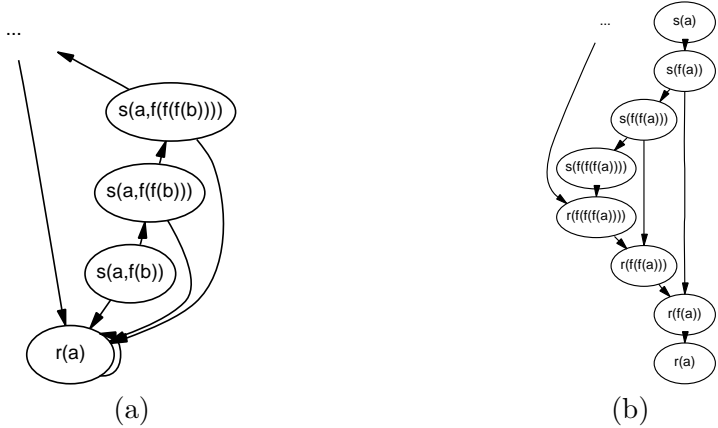

$$(a) \qquad\qquad\qquad (b)$$

Figure 6. Part of the *directly influence* relation of (a) the first and (b) the second ill-defined Bayesian logic programs of example 4.10.

## 5. Query-Answering Procedure

In this section, we show how to answer probabilistic queries to a Bayesian logic program. A probabilistic query to a Bayesian logic programs is with respect to Prolog defined as follows:

**Definition 5.1** (*probabilistic query*). *A probabilistic query to a Bayesian logic program $B$ is an expression of the form* ?- Q₁,...,Qₙ | E₁ = e₁,...,Eₘ = eₘ *with $n > 0$, $m \geq 0$. It asks for the conditional density $\mathbf{p}(Q_1,...,Q_n \mid E_1 = e_1,...,E_m = e_m)$ of the query variables $Q_1,...,Q_n$ where $\{Q_1,...,Q_n, E_1,...,E_m\} \subset \mathrm{HB}(B)$. A query with $m = 0$ is called evidence-free.*

The definition generalizes the inference problem for Bayesian networks, see definition 2.2. Due to Theorem 4.9, we say that an answer is defined if and only if $\{Q_1,...,Q_n, E_1,...,E_m\} \subset \mathrm{LH}(B)$. To answer a query we adapt the two step strategy of *knowledge-based model construction* approaches: (see e.g. [BGW94,Had99]): first, we construct a Bayesian network $N$ and, second, we apply a Bayesian network inference algorithm on $N$ in order to answer the query. We first assume that the answer is defined and will come back to the more general case in Section 5.3.

A naive approach would be to explicitly build the Bayesian network representing the *direct influenced by* relation over the proper random variables. Because the resulting Bayesian network may be infinity, this is impossible. The way out of the problem is to use *support networks*. The notion of support network is due to Ngo and Haddawy [NH97] but we will adapt it for our purposes.

**Definition 5.2** (*support network*). *Let $B$ be a Bayesian logic program, $N$ its (possibly infinite) Bayesian network and $X_1,...,X_m$, $m > 0$, nodes of $N$. The* support network $N(X_1, \ldots, X_m)$ of $X_1, \ldots, X_m$ is the subnetwork of $N$ which consists of the nodes

$$\{Y \in \mathrm{LH}(B) \mid Y \text{ influences some } X_i,\ 1 \le i \le m\} \tag{5.1}$$

and the edges of $N$ which connect only nodes in $N(X_1, \ldots, X_m)$.

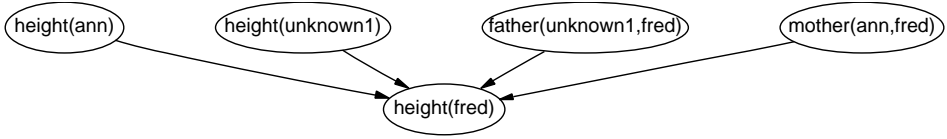

Figure 7. The support network $N(height(fred))$ with respect to the *height* Bayesian logic program.

The support network $N(height(fred))$ with respect to our running example is shown in Figure 7. We will now prove that the support network is sufficient to compute any conditional probability density involving only random variables of the support network.

**Theorem 5.3.** *Let $N$ be a possibly infinite Bayesian network, $Q_1, \ldots, Q_n$, $n > 0$, nodes of $N$ and $\mathbf{E} = \mathbf{e}$, $\mathbf{E} \subset N$. The computation of $\mathbf{p}(Q_1, \ldots, Q_n \mid \mathbf{E} = \mathbf{e})$ does not depend on any node $X$ of $N$ which is not a member of the support network $N(\{Q_1, \ldots, Q_n\} \cup \mathbf{E})$.*

*Proof.* This proof can be skipped without loss of continuity. Background material and definitions from probability theory are introduced in Section A in the appendix.

In order to prove the theorem we only have to show that $N(\{Q_1, \ldots, Q_n\} \cup \mathbf{E})$ is sufficient to compute $\mathbf{p}(X_1, \ldots, X_l)$ for any set $\{X_1, \ldots, X_l\}$, $l > 0$, of random variables in $N(\{Q_1, \ldots, Q_n\} \cup \mathbf{E})$. The theorem follows then from the definition of conditional probability density:

$$\mathbf{p}(Q_1, \ldots, Q_n \mid \mathbf{E} = \mathbf{e}) = \frac{\mathbf{p}(Q_1, \ldots, Q_n, \mathbf{E} = \mathbf{e})}{p(\mathbf{E} = \mathbf{e})}$$

We proceed in a similar way to the proof of theorem 4.9. Let $\pi$ be a total order of the nodes in $N$. Let $T$ be a (non-empty) index set, $\mathcal{H}(T)$ be the set of all non-empty, finite subsets of $T$, and $(A_n)_{n \in T}$ be the sequence of random variables in $N$ in ascending order to $\pi$. Analogously to the proof of theorem 4.9 we can prove by induction that $N$ specifies a projective family of probability measure. Now, the set $\{Q_1, \ldots, Q_n\} \cup \mathbf{E}$ corresponds to $\mathbf{A}(\mathbf{H})$ for some $H \in \mathcal{H}(T)$, and the set $\{X_1, \ldots, X_l\}$ corresponds to $\mathbf{A}(\mathbf{L})$ for some $L \in \mathcal{H}(T)$. In order to compute $\mathbf{p}(X_1, \ldots, X_l)$ we consider the completion

$$\mathbf{C}(\mathbf{L}) = \{D \in N \mid D \text{ influences some } X_i \in \mathbf{A}(\mathbf{L})\}$$

of $\mathbf{A}(\mathbf{L})$ (resp. $\mathbf{C}(\mathbf{H})$ of $\mathbf{A}(\mathbf{H})$). The set $\mathbf{C}(\mathbf{H})$ equals per definitionem the set of nodes of $N(\{Q_1, \ldots, Q_n\} \cup \mathbf{E})$. Therefore, we have $\mathbf{A}(\mathbf{L}) \subset \mathbf{C}(\mathbf{H})$ and, hence, $\mathbf{C}(\mathbf{L}) \subset \mathbf{C}(\mathbf{H})$. As in the proof of theorem 4.9, the probability densities over $\mathbf{C}(\mathbf{H})$ (resp. $\mathbf{C}(\mathbf{L})$) are specified by a unique Bayesian network $N(H)$ (resp. $N(L)$). It consists of all random variables in $\mathbf{C}(\mathbf{H})$ (resp. $\mathbf{C}(\mathbf{L})$) and of all edges between nodes in $N$ which are random variables in $\mathbf{C}(\mathbf{H})$ (resp. $\mathbf{C}(\mathbf{L})$). Since $N$ specifies a projective family of probability measures, $N(L)$ is a subnetwork of $N(H)$. That means the computation of $\mathbf{p}(X_1, \ldots, X_l)$ only depends on nodes and edges in $N(H)$. But $N(H)$ is per definition the support network $N(\{Q_1, \ldots, Q_n\} \cup \mathbf{E})$. The theorem is proven. $\square$

Thus, we can answer the probabilistic query `?- Q_1,...,Q_n | E_1 = e_1,...,E_m = e_m` using the support network $N(\{Q_1, \ldots, Q_n\} \cup \{E_1, \ldots, E_m\})$. The next two sections deal with computing support networks. They rely on the following property of support networks:

**Proposition 5.4.** *Let $B$ be a Bayesian logic program, $N$ its (possibly infinite) Bayesian network and $X_1, \ldots, X_m$, $m > 0$, nodes of $N$. The support network $N(X_1, \ldots, X_m)$ is the graph union $G$ of all single support networks $N(X_i)$.*

*Proof.* First, we show that $N(X_1, \ldots, X_m)$ and $G$ have the same set of nodes. The support network $N(X_1, \ldots, X_m)$ has per definitionem a node $A$ if and only if $A$ is influencing a $X_i \in \{X_1, \ldots, X_m\}$. But, $A$ is influencing a $X_i \in \{X_1, \ldots, X_m\}$ if and only if $A$ is a node in $N(X_i)$, i.e. $A$ is a node in $G$.

Now, we prove that $N(X_1, \ldots, X_m)$ and $G$ have the same set of edges. A support network $N(X_1, \ldots, X_m)$ has an edge $E$ from a node $A_i$ to a node $A_j$ if and only if (1) both nodes, $A_i$ and $A_j$ are influencing a $X_k \in \{X_1, \ldots, X_m\}$ and are therefore in $N(X_k)$, and (2) the edge $E$ is in $N$. Per definitionem of a support network this is if an only if the edge $E$ is in $N(X_k)$, i.e. $E$ is an edge in $G$. $\square$

### 5.1. Evidence-free Probabilistic Queries

We first restrict ourselves to evidence-free queries of the form `?- Q_1`.

**Example 5.5.** Given the Bayesian logic program of example 3.10, the probabilistic query `?- height(fred)` asks for the densities $\mathbf{p}(height(fred))$. The information about *john* stated in $B$ is irrelevant, because it does not appear in the support network $N(height(fred))$ shown in Figure 7. $\square$

But how do we build the support network?

#### 5.1.1. AND/OR Trees

The usual execution model of logic programs relies on the notion of SLD trees (see e.g. [Llo89,SS86]). For our purposes AND/OR trees [Nil71,KK71,VM75, Nil86] are more suitable because they allow us, as we will see, to combine the probabilistic with the logical computations. AND/OR trees have a long history in the AI community and we adapt them here for our purposes.

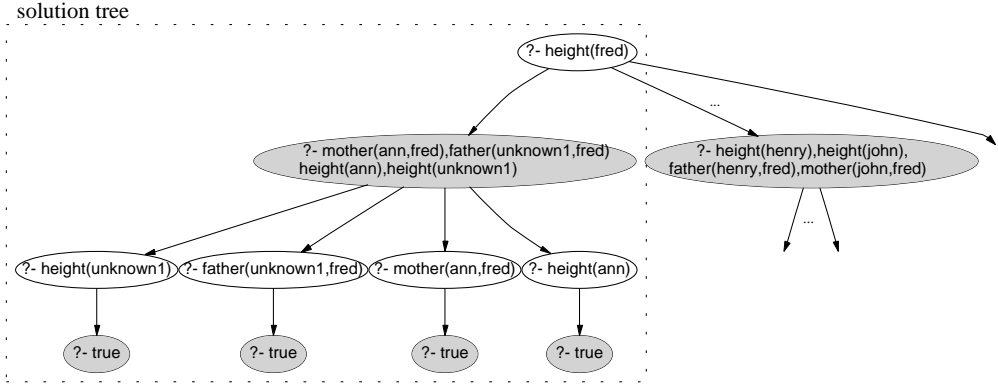

Figure 8. The AND/OR tree $T(height(fred))$ according to example 5.8. Unfilled ovals represent *or* nodes, whereas *and* nodes are represented by filled ovals. The dotted box indicates the solution graph $S(height(fred))$, i.e. all unsolved nodes together with their in- and outgoing edges; in particular all infinite paths are removed.

**Definition 5.6** *(AND/OR tree). Let $B$ be a Bayesian logic program and* `?- Q` *an evidence-free probabilistic query. The AND/OR tree $T(Q)$ of the query given $B$ is a tree whose nodes are divided into two disjunctive sets, the set of and nodes and the set of or nodes. Each node contains a conjunction of ground atoms. The nodes* `?- A₁,...,?- Aₙ` *constitute all children of an and node* `?- A₁,...,Aₙ`*. An or node* `?- A` *has a child* `?- (A₁,...,Aₙ)θ` *if a Bayesian definite clause $A' \mid A_1,...,A_n$ in $B$ and a substitution $\theta$ exist, such that $\theta$ grounds the clause and $A'\theta = A$. The only node* `?- Q` *having no predecessors is called the root node and it is always an or node.*

In describing AND/OR trees we shall continue to use terms like *parent* nodes, *successor* nodes, *paths* etc. with the obvious meaning. While constructing an AND/OR tree we interpret a ground fact `A` as a ground clause `A | true` where the symbol `true` is a built-in with the usual meaning. Our objective for using an AND/OR tree $T(Q)$ is to show that $Q$ is solved, i.e. $Q \in \mathrm{LH}(B)$.

**Definition 5.7** *(solved). An and node containing* `?- true` *is solved. Any other and node is solved if it has at least one child and all of its children are solved. An or node is solved if at least one of its children is solved.*

**Example 5.8.** Consider again the Bayesian logic program of example 3.10. The AND/OR tree $T(height(fred))$ of Figure 8 states that `height(fred)` is solved because the *and* node

```
?- mother(ann, fred), father(unknown1,fred), height(ann), height(unknown1)
```

is solved. This node in turn is solved because `?- father(unknown1,fred)`, `?- mother(ann, fred)`, `?- height(ann)` and `?- height(unknown1)` are all solved. □

**Definition 5.9** *(solution tree [Nil71]).* *Let $T(Q)$ be an AND/OR tree. The solution tree $S(Q)$ of $T(Q)$ is the maximally finite subtree of solved nodes of $T(Q)$.*

A solution tree partly represents the *immediate consequence* operator and hence a subset of the *directly influenced by* relation over $\mathrm{LH}(B)$. This is due to the following property:

**Property 5.10.** *There exists an edge from a solved or node labeled $A$ to a solved and node labeled $C$ if and only if there exists a ground instance $A \mid C$ of a clause in $B$ such that $\{A\} \cup C \subset \mathrm{LH}(B)$.*

All *or* nodes of a solution tree correspond to proper random variables; we call the set of random variables corresponding to the nodes *relevant* with respect to the given Bayesian logic program and probabilistic query. A solution tree $S(Q)$ does not only encode $Q \in \mathrm{LH}(B)$ and its set of relevant random variables, it encodes also all ways of proving that its *or* nodes are solved. It follows that the solution tree is unique. This makes it possible to represent the solution tree in a more compact and more suitable way: its *collapsed* version. All nodes containing the same query are merged. We call the resulting graph the *solution graph $S(Q)$*. The solution graph is a more suitable representation because there exists a one-to-one mapping between the *or* nodes of $S(Q)$ and the relevant random variables, so that the following properties hold

**Property 5.11.** *There exists an edge $e$ in $S(Q)$ going from an or node $O$ to an and node $A$ if and only if there exists a clause $C \in B$, such that $O \mid A$ is a ground instance of $C$ and $\{O\} \cup A \subset \mathrm{LH}(B)$.*

**Property 5.12.** *Let $O$ be an or node $O$ in $S(Q)$ and $X \in \mathrm{LH}(B)$ its corresponding random variable. Furthermore, let $\{X_1, \ldots, X_n\} \subset \mathrm{LH}(B)$ be the random variables corresponding to the grandchildren of $O$ in $S(Q)$. Then $\mathbf{Pa}(X) = \{X_1, \ldots, X_n\}$.*

Hence, we can augment the edge $e$ with the conditional probability densities $cpd(C)$, denoted as $cpd(e)$.

**Example 5.13.** Figure 9 shows the solution graph of the probabilistic query `?- height(fred)` augmented with conditional probability densities. □

Now, we have everything to prove the main theorem of this subsection.

**Theorem 5.14.** *Let $B$ be a well-defined Bayesian logic program and `?- Q` an evidence-free probabilistic query with $Q \in \mathrm{LH}(B)$. The support network $N(Q)$ is the result of performing for each or node $A$ (over predicate $r$) in the solution graph $S(Q)$ the following steps:*

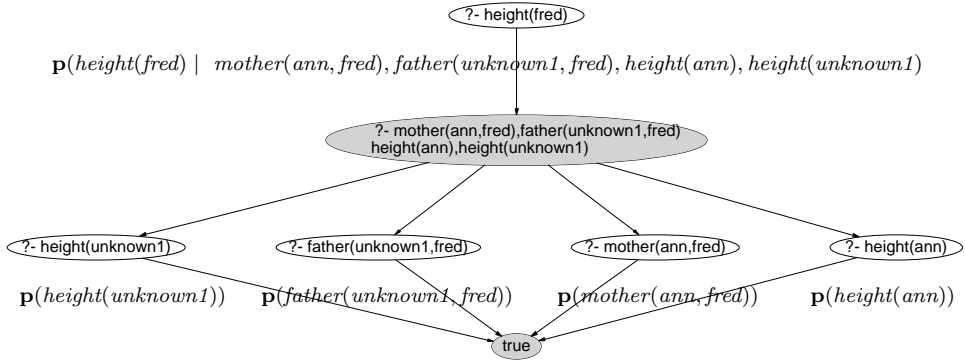

PSfrag replacements

Figure 9. The (collapsed) solution graph of `height(fred)` augmented with probabilistic information: The conditional probability densities $\mathbf{p}(O \mid A)$ are associated to each edge from an *or* node $O$ to an *and* node $A$. The densities $\mathbf{p}(O \mid A)$ equal *cpd(c)*, the densities associated to the Bayesian clause corresponding to the edge.

1. *Compute the combined conditional probability densities of $A$:*

   $$cpd(A) = comb(r)\{cpd(e) \mid e \text{ is an outgoing edge of } A\}.$$

2. *Associate cpd(A) to the node $A$.*

3. *Remove all children of $A$ in $N$ and their in- and outgoing edges.*

4. *Insert a directed edge from each formerly grandchild of $A$, on which cpd(A) is conditionalized, to $A$ itself.*

*The resulting augmented graph is $N(Q)$.*

*Proof.* Let $B$ be a well-defined Bayesian logic program and `?- Q` a probabilistic query to $B$ with $Q \in \mathrm{LH}(B)$. Remember that the solution graph $S(Q)$ is given in its collapsed version and that it is augmented with the corresponding probability distributions. It is clear that the described transformation on $S(Q)$ yields a Bayesian network $N$ whose dependency structure coincides with a subset of the *directly influenced by* relation over $\mathrm{LH}(B)$ and which encodes the joint probability density. But we have to prove that $N$ is a support network of $Q$. It is trivial that $N$ fullfils the conditions 1, 2 and 3 of definition 5.2. What remains to be proven is that $N$ is of minimal size. Let us assume that $N$ is not minimal, i.e. that a Bayesian network $N'$ exists which fullfils the conditions 1, 2 and 3 but is of smaller size. Because of condition 3 we can assume that $N$ has at least one node $U$ which is not a node of $N'$. Due to condition 1, $U$ is not the root node $Q$ of $N$. So, let $U$ be any other node except for the root node. We can follow a path from $U$ through its children and must come to a node $U'$ in $N'$ (at least the root node, which is identical in $N$ and $N'$). This node has $\mathbf{Pa}(U') \not\subset N'$. Thus $U'$ violates condition 2. This contradicts our assumption that $N'$ is a smaller Bayesian network fullfilling the conditions 1, 2 and 3. $\qquad\square$

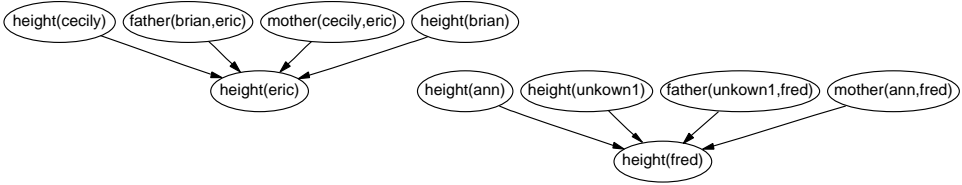

Figure 10. The support network of the query `?- height(fred) | height(eric)` of example 5.15.

### 5.2. Probabilistic Queries With Evidence

So far, we have restrict ourselves to evidence-free queries of the form `?-Q₁`. The extension to the more general case concerns queries of the form

$$\text{?-}\mathtt{Q_n},\dots,\mathtt{Q_n} \mid \mathtt{E_1 = e_1},\dots,\mathtt{E_m = e_m}$$

with $n > 0$, $m > 0$. Due to theorem 5.3 we have to build the support network $N(\{Q_1,\dots,Q_n,E_1,\dots,E_m\})$. Due to proposition 5.4 this is the graph union of all single support networks $N(Q_1),\dots,N(Q_n),N(E_1),\dots,N(E_m)$. Consequently, the procedures and claims about evidence-free queries can easily be adapted.

### 5.3. Characteristics of the Query-Answering Procedure

Here, we study the minimality of the support network as well as the soundness of the purposed inference procedure.

Let us first point out that the support network is not minimal with respect to given instance of the inference problem.

**Example 5.15.** Figure 10 shows the support network for

$$\text{?- height(fred) | height(eric).}$$

It consists of two connected components, one for `height(fred)` and one for `height(eric)`, of which the latter one is redundant. □

This is related to a well-known problem in Bayesian networks: which nodes of a given Bayesian network are relevant to compute a desired density? Various answers exists in the literature, e.g. Geiger et. al. [GVP90] or Shachter's Bayes-Ball [Sha98], and all of them can be applied to the constructed support network. One refinement to our algorithm is due to Ngo and Haddawy [NH97]: if no (undirected) path between the query variable and an evidence variable exists, then the support network of that evidence variable can be deleted.

If one of the random variables occurring in a query is not proper then the solution graph of that variable is empty and, hence, no density is specified. The answer to the query is undefined. Furthermore, if the Bayesian logic program is ill-defined, e.g. due to infinite branching factors or an infinite path then the procedure would not terminate, too.

**Definition 5.16** *(well-defined probabilistic query). A probabilistic query* $\mathtt{Q_1, \ldots, Q_n \mid E_1 = e_1, \ldots, E_m = e_m}$ *is well-defined, if the AND/OR graphs of all variables* $Q_1, \ldots, Q_n, E_1, \ldots, E_m$ *are finite.*

The definition resembles the definition of well-defined Bayesian logic programs in compliance with computability. Remember that our query-answering procedure follows the *knowledge-based model construction* approach. It consists of two phases: (1) construct the support network of the given probabilistic query and (2) apply a Bayesian network inference algorithm to the support network. Thus, we have to guarantee that the construction of the support network terminates. The AND/OR tree of each random variables in a well-defined query is finite. The construction of the support network terminates in finite time. Thus, our proposed query-answering procedure is sound for well-defined probabilistic queries.

## 6. Interpreting Bayesian Logic Programs

Though the previous section describes a query-answering procedure, the question of how to compute the solution graphs themselves is still open. This section will show, that Bayesian logic programs can be interpreted using a common meta interpreter written in Prolog. This in turn gives us an efficient and particularly practical approach for constructing support networks.

### 6.1. The "missing" link between solution graphs and SLD trees

To come up with such a meta interpreter we analyze the relation of solution graphs to SLD trees [Llo89,SS86]. The relation was already investigated in the early days of Artificial Intelligence by Kowalski and Kuehner [KK71, page 250]. But for reasons of self-containedness we will discuss it here.

The SLD tree represents all possible series of application of the SLD resolution inference rule on a goal (cf. [Llo89]) graphically. A *goal* is an expression of the form *?- $G_1, \ldots, G_n$* where all $G_i$'s are Bayesian atoms. For Bayesian logic programs, we have:

**Definition 6.1** *(SLD tree). Let B be a Bayesian logic program and ?- G a a goal. An SLD tree for* $B \cup \{G\}$ *fullfils the following conditions:*

1. *Each node A of the tree is a (possibly empty) goal. The set of atoms of A is denoted by atoms(A).*

2. *Let ?- $A_1, \ldots, A_m, \ldots, A_k$ ($k \geq 1$) be a node of the tree where $A_m$ is the selected atom. Then, this node has for each clause $A := B_1, \ldots, B_n$ in B, such that $A_m$ and A unify with the most general unifier (MGU) $\theta$, a successor node*

$$?\text{-} (A_1, \ldots, A_{m-1}, B_1, \ldots, B_n, A_{m+1}, \ldots, A_k)\theta.$$

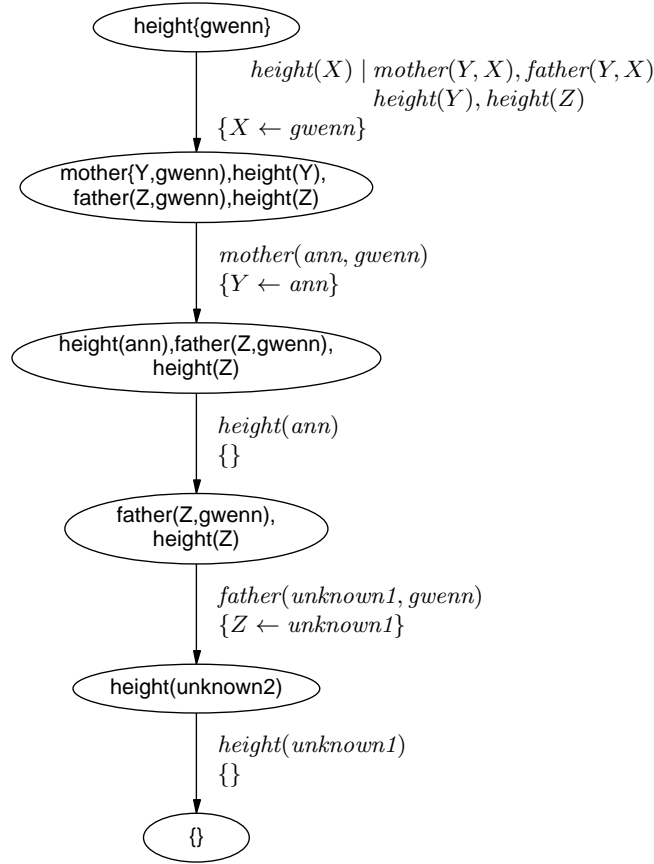

Figure 11. The SLD tree of the goal `?- height(gwenn)`. The edges are labeled with the applied MGUs and clauses. The corresponding solution graph is shown in Figure 9.

3. *Nodes representing the empty clause have no successor.*

Thus, a single edge corresponds to a single application of the resolution inference rule which could be seen as a kind of Modus Ponens in the case of definite clause programs [RN95]. In describing SLD trees we shall continue to use terms like *parent* nodes, *successor* nodes, *paths* etc. giving them the obvious meaning. A *successful* path is a path starting at the root and leading to a terminal node representing the empty clause. The (finite) SLD tree of the goal `?- height(gwenn)` is shown in Figure 11. We label each edge $e$ of an SLD tree with the MGU $\theta(e)$ and the clause $C(e)$ used in the corresponding application of the resolution inference rule. Let $S$ be an SLD tree of a (well-defined) Bayesian logic program $B$ and `?-` $A_1, \ldots, A_k$ a goal $G$. Furthermore, let $P_1, \ldots, P_n$ be the nodes along a *successful* path $P$ of length $n$ in $S$ and $e_i$ the edge pointing from $P_i$ to $P_{i+1}$. It is well-known (see e.g. [Llo89]) that for range-restricted definite clauses programs

(such as Bayesian logic programs)

$$\bigcup_{j=1}^{k} (A_j \theta_{G,P}) \subseteq LH(B),$$

where $\theta_{G,P}$ is the substitution $\theta_{G,P} = \theta(e_1) \dots \theta(e_{n-1})$. The substitution $\theta_{G,P}$ is called the *answer substitution*. Because every node $P_i$ in an SLD tree is itself a goal the same is true for all nodes along $P$, i.e. $\bigcup_{A \in atoms(P_i)} (A\theta_{P_i,P}) \subseteq LH(B)$. Thus, each grounded node $P_i\theta_{G,P}$ consists of proper random variables. The ground clause $C(e_j)\theta_{G,P}$ corresponds to an edge from the *or* node $head(C(e_j))\theta_{G,P}$ to the *and* node $body(C(e_j))\theta_{G,P}$ in a solution graph. According to the definition of AND/OR trees, all children of the *and* node $body(C(e_j)\theta_{G,P})$ are determined by the node itself. Therefore, if we require all $A_i$ to be ground atoms then we can construct the union of the solution graphs of $A_1, \dots, A_k$ from the set of all grounded successful paths.

## 6.2. An implementation

Having established the link between SLD trees and solution graphs of a Bayesian logic program it is easy to interpret Bayesian logic programs. E.g. one could adapt a backward chaining algorithm presented in [RN95, page 275] as done in the procedures SUPPORTNETWORK, SLD-TREE and COMPUTESUPPORTNET-WORK.

The procedure SUPPORTNETWORK establishes the overall flow. After initializing the support network $N$ and the SLD tree *Tree* to be empty, it computes both of them calling first SLD-TREE and then COMPUTESUPPORTNETWORK. At the end, one could prune the support network $N$, although we will not investigate this.

The backward-chaining algorithm SLD-TREE works by first checking to see if any Bayesian ground fact unifies with the query $Q$. If so, corresponding edges are inserted into the SLD tree *Tree*. It then finds all Bayesian clauses whose head unifies with the query $Q$, and tries to prove the bodies of those Bayesian clauses, also by backward-chaining. SLD-TREE processes the body of a selected Bayesian clause atom by atom, building up the whole SLD tree *Tree*. Remember that each selected Bayesian clause together with the query and the MGU specifies a particular set of edges in the SLD tree *Tree*. The clause, the MGU and the corresponding associated conditional probability densities are stored together with the edge in *Tree*.

COMPUTESUPPORTNETWORK inspects all successful paths $P$ in *Tree* in turn and builds up the corresponding support network $N$. It works by inserting for a ground clause $C(e)\Theta$ (gathered from the informations associated to an edge $e$ in $P$) all corresponding nodes and edges in the support network $N$. After building this "uncombined" version of the support network $N$, it traverses $N$ and applies the corresponding combining rules on each node. This may delete some of the edges.

A slightly modified and naive implementation of these procedures in Prolog can be found in the Appendix. It builds on a Prolog meta interpreter. Various types of such meta interpreters relying on the SLD tree exists (see e.g. [SS86,Bra86]). We adapted a simple one relying on *depth-first search*.

---

**Data** : $B$, a well-defined Bayesian logic program; $Q$, a query variable.
**Result**: $N$, the support network of `?- Q`.

$N \leftarrow$ EMPTYBAYESIANNETWORK;
*Tree* $\leftarrow$ EMPTYSLDTREE;
SLD-TREE($B, Q,$ *Tree*);
COMPUTESUPPORTNETWORK(*Tree*, $N$);
$N \leftarrow$ PRUNE($N$);

---

**Algorithm 1:** SUPPORTNETWORK($B, Q, N$)

## 7. Examples of Bayesian Logic Programs

In this section, we illustrate the representational power and elegance of Bayesian logic programs by demonstrating that Bayesian network, definite clause programs (as in "pure" Prolog), hidden Markov models and dynamic Bayesian networks can straightforwardly be encoded as Bayesian logic programs. Furthermore, we give examples of Bayesian logic programs involving structured terms (and having a countably infinite least Herbrand model). But before doing so, let us put Bayesian logic programs in a wider context. The proof of Theorem 4.9 makes a more abstract interpretation of Bayesian logic programs possible. Given a total order on the least Herbrand model a Bayesian logic program represents a discrete-time stochastic processes (cf. the appendix). If we see the total order as a time line, then the state of a single proper random variable could depend on all variables in the past. Such processes are called *infinite-memory* This places Bayesian logic programs in a wider context of what Cowell et. al. call *highly structured stochastic systems* (cf. [CDLS99]). Well-known probabilistic frameworks such as dynamic Bayesian networks, hidden Markov models or Kalman filters are special cases of such highly structured stochastic systems.

---

**Data** : *Tree*, an SLD tree.

**Result**: $N$, a support network.

**foreach** *successful path $P \in$ Tree where $e_1, \ldots, e_n$ are the edges of $P$ starting at the root node* **do**

    $\theta \leftarrow$ composition of $\theta(e_1), \ldots, \theta(e_n)$;

    **for** $i = n$ ***down to*** 1 **do**

        insert node $head(C(e_i)\Theta)$ into $N$;

        **if** *$C(e_i)\Theta$ is not a ground fact* **then**

            **foreach** *ground atom $b \in body(C(e_i)\Theta)$* **do**

                insert node $b$ into $N$;

                insert an edge from $b$ to $head(C(e_i)\Theta)$ into $N$;

            **end**

        **end**

        Store $cpd(C(e_i))$ at node $head(C(e_i)\Theta)$ ;

    **end**

**end**

Apply to each node in $N$ over a predicate $r$ the corresponding combining rule $comb(r)$;

/* Note that this may delete some edges in $N$ */;

---

**Algorithm 2:** ComputeSupportNetwork(*Tree*, $N$)

### 7.1. Bayesian networks

Section 3.1 has already shown that every Bayesian network directly translates to a propositional Bayesian logic program. Here, we give a further illustration. Consider the famous example due to Judea Pearl about burglary alarms at home (cf. e.g. [RN95]). It translates to

```
burglary.
earthquake.
alarm      | burglary, earthquake.
johncalls  | alarm.
marycalls  | alarm.
```

where the associated conditional probability densities are identical to the densities of the Bayesian network.

### 7.2. Definite Clause Logic

Another interesting subclass of Bayesian logic programs are definite clause programs, i.e. "pure" Prolog programs. They give us the power of "pure" Prolog to

| $\mathbf{P}(A \mid A_1, \ldots, A_n)$ | | | | | |
|---|---|---|---|---|---|
| *true* | *false* | $A_1$ | $A_2$ | $\ldots$ | $A_n$ |
| 1.0 | 0.0 | *true* | *true* | | *true* |
| 0.0 | 1.0 | *false* | *true* | | *true* |
| $\vdots$ | $\vdots$ | $\vdots$ | $\vdots$ | | $\vdots$ |
| 0.0 | 1.0 | *false* | *false* | | *false* |

Table 2

The conditional probability densities associated to a logical Bayesian clause `A | A₁,...,Aₙ`.

model deterministic knowledge within the framework of Bayesian logic programs. Let us start with a simple example. The logic program *parents*

```
father(jef,paul).
mother(an,paul).
parent(X,Y)      :- father(X,Y).
parent(X,Y)      :- mother(X,Y).
```

defines the parent relation in terms of `father` and `mother`.

**Example 7.1.** We model this with the Bayesian logic program

```
father(jef,paul).
mother(an,paul).
parent(X,Y)   |   father(X,Y).
parent(X,Y)   |   mother(X,Y).
```

Each predicate has as domain $\{true, false\}$ and max as combining rule. To each clause $C$ we associate $cpd(C)$ as in Table 2. It is easy to compute $p(father(jef, paul) = true) = 1.0$ and $p(parent(jef, paul) = false) = 0.0$.  $\square$

Choosing other associated conditional probability densities one could easily implement some forms of negation such as explicit negation. The example generalizes to the following theorem.

**Theorem 7.2.** *Let $L$ be a definite clause program, such that the solution graph of each ground atom in $\mathrm{LH}(L)$ is finite. Let $B$ be the Bayesian logic program which is the result of applying the transformation of example 7.1 on $L$. Furthermore, let $G$ be a Bayesian ground atom. Then, $B$ specifies $P(G = true) = 1.0$ if and only if the logical ground atom of $G \in \mathrm{LH}(L)$.*

*Proof.* Let $L$ be a definite clause program, such that the solution graph of each ground atom in $\mathrm{LH}(L)$ is finite, and $B$ the Bayesian logic program which is the result of applying the transformation of example 7.1 on $L$.

"$\Rightarrow$": The program $B$ only specifies a density over $G$ when $G \in \mathrm{LH}(B)$. Because $L$ is the corresponding logic program of $B$, $G \in \mathrm{LH}(L)$ holds.

"⇐": Let $G \in \mathrm{LH}(L)$. Because $L$ is the corresponding logic program of $B$, it follows $G \in \mathrm{LH}(B)$. Furthermore, the solution graph of $G$ represents all possible proofs of $G$. An induction over the number of nodes of the solution graph utilizing the max combining rule proves $P(G = true) = 1.0$. □

The restrictions of theorem 7.2 are quite strong. First, the logic programs must fullfill the conditions of a well-defined Bayesian logic program, i.e. the solution graph of any provable ground atom must be finite. Second, the answer to a probabilistic query of ground atoms not occurring in the least Herbrand model is undefined. E.g. the *parents* logic program does not entails `parent(jo,henry)`. The solution graph of `parent(jo,henry)` is empty. In the second case, we refer to Remark 5.3 and theorem 7.2. They justify to interpret this "null-value" as $P(Q = false) = 1.0$. But still the first restriction is very strong. Many logic programs (see e.g. example 4.10) do not fit theorem 7.2. Some provable ground atoms have infinitely many proofs. We stipulate in theorem 4.9 a finite branching factor of the solution graphs by virtue of finite computations, especially of the combining rules. But if we modify the definition 3.6 of combining rules to allow for infinite input sets, then the declarative semantics of Bayesian logic programs could be extended to cover logic programs having solution graphs of finite depth but infinite branching factors (see example 4.10). Under the modified declarative semantics we could formulate the following theorem:

**Theorem 7.3.** *Let $L$ be a definite clause program and $B$ the corresponding logical Bayesian logic program. Then, $B$ specifies $P(G = true) = 1.0$, $G \in \mathrm{LH}(B)$, if and only if the corresponding logical ground atom of $G$ is a member of $\mathrm{LH}(L)$.*

This gives us the possibility to effectively model logical knowledge within Bayesian logic programs.

### 7.3. Structured terms

Consider the following well-defined Bayesian logic program (the associated conditional probability densities are not important):

```
even(0).
even(s(X)) | odd(X).
odd(s(X))  | even(X).
```

It is easy to see that we can compute the answer to any probabilistic queries. such as `odd(s(0)) | even(s(s(0))) = true`. We have to compute the support network, which is $even(0) \rightarrow odd(s(0)) \rightarrow even(s(s(0)))$, feed it into a Bayesian network engine and compute the answer. One can easily see that despite the presence of structured terms and an infinite number of random variables, the required computations are finite.

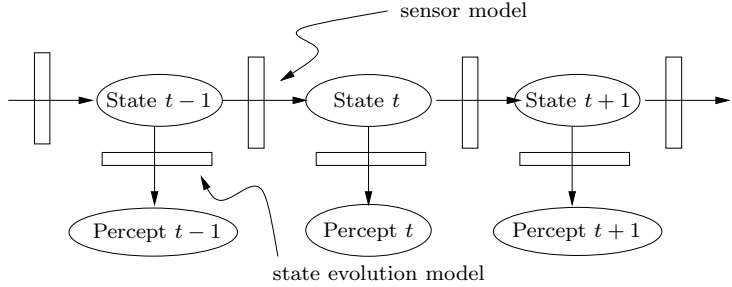

Figure 12. The generic structure of a dynamic Bayesian network (see [KKR95]). The associated conditional probability densities quantifying the transition probabilities between states are called the *state evolution model* and the conditional probability densities describing the observations which can result from a state form the *sensor model*.

### 7.4. Dynamic Bayesian networks

A framework covering both Bayesian networks and hidden Markov models is that of dynamic Bayesian networks [Kjæ95]. The fundamental idea is to divide time into *time slices*, each representing a snapshot of the evolving temporal process. These states of the world at a specific time point are described using Bayesian subnetworks over a finite sets of random variables. According to Kanazawa et. al. [KKR95] "a dynamic Bayesian networks consists of a sequence of time slices where nodes within time slice $t$ are connected to nodes in time slice $t + 1$ as well as to other nodes within slice $t$." Hence, each time slice $t$ is a Bayesian subnetwork. The generic structure of a dynamic Bayesian network is shown in Figure 12. If one assumes that the associated conditional probability densities do not vary over time (a common assumption, see e.g. [KKR95]), one can show that dynamic Bayesian networks can be represented using Bayesian logic programs by combining the ideas of Section 7.1 and Section 7.3. We encode the starting Bayesian network of time slice 0 using Bayesian atoms having 0 in the last argument. The connection between a time slice $t$ and a time slice $t+$ are modeled using clauses involving the term `succ(T)` in the head and the term `T` in the body. Therefore, Bayesian logic programs clearly generalize dynamic Bayesian networks. However, the structure expressible with Bayesian logic programs is much more flexible than that of dynamic Bayesian networks. E.g. the structure or the random variables of the time slices can vary over time. Hence, Bayesian logic programs could better represent a particular situation in time. For a discussion on why this is important we refer to the work of Glesner and Koller [GK95].

## 8. Related Work

Bayesian logic programs are related to all combinations of first order logic with probability theory. However, we will focus on first order extensions of Bayesian networks only. Other works such as the one by Ng and Subrahmanian [NS92], who introduced a probabilistic characterization of logic programming, by Sato [Sat95], or by Cussens and Muggleton on stochastic logic programs [Mug96,Cus99b] will not be treated in this paper. We refer for surveys to Parson's article [Par96] and Section 3 of Cussen's paper [Cus99a]. This is in line with Halpern's [Hal89] analysis of first order logics of probability. Halpern introduced two probabilistic structures. A structure of *type I* represents the *degree of belief* of an agent. Bayesian networks and Bayesian logic programs are examples of this type of structure. A structure of *type II* represents statistical knowledge. One can e.g. express the probability that a randomly chosen object has some property. Stochastic logic programs are an example of this second type of structure. In contrast to Bayesian logic programs, probabilities in stochastic logic programs are defined directly on the proofs of atomic formulas.

Bayesian logic programs are motivated and inspired by the formalisms discussed in [Poo93,Had94,NH97,Jae97,FGKP99,Kol99]. They are most closely related to Ngo and Haddawy's *knowledge-based model construction* framework of probabilistic logic programs [NH97] (cf. Section 8.1). The idea of associating conditional probability densities to clauses is also proposed by Fabian and Lambert [FL98], though they view a ground atom as a random variable over {*true, false*} and give a quite different declarative semantics. E.g. they do not have the concept of a combining rule but instead use a kind of backtracking mechanism to cope with the situation where one atom can be proven in different ways.

We will now more closely investigate the relationship of Bayesian logic programs to Ngo and Haddawy's probabilistic logic programs [NH97], Pooles's probabilistic Horn abduction [Poo93], Koller et. al.'s probabilistic relational models [Kol99,FGKP99] and Jaeger's relational Bayesian networks [Jae97].

### 8.1. Probabilistic logic programs

Probabilistic logic programs [NH95,NH97] follow the *knowledge-based model construction* technique and adapt as Bayesian logic programs the concept of least Herbrand model to specify the relevant random variables. A query-answering procedure exists and is based on SLD resolution. An example probabilistic logic

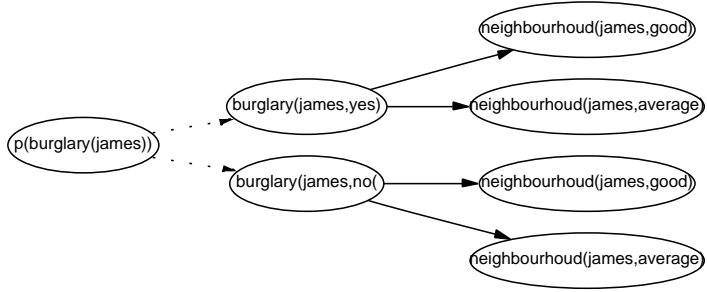

Figure 13. The (grounded) SLD trees built to compute $p(burglary(\texttt{james}))$ within probabilistic logic programs. The dotted edges indicates that the two SLD trees are computed in order to compute $p(burglary(\texttt{james}))$.

program (following [NH97]) is

$$P(neighbourhood(X, average)) = 0.4$$
$$P(neighbourhood(X, good)) = 0.3$$
$$P(burglary(X, yes) \mid neighbourhood(X, average)) = 0.4$$
$$P(burglary(X, no) \mid neighbourhood(X, good)) = 0.7$$

It consists of four so-called probabilistic sentences. Each such sentence quantifies a probabilistic dependency among random variables, e.g. the aposteriori probability of a burglary in the house of person $X$ given that $X$ has a good neighbourhood is 0.7. Even this simple example shows five main differences between probabilistic logic-programs and Bayesian logic programs.

First, in probabilistic logic programs ground atoms $g(t_1, \ldots, t_{n-1}, t_n)$ correspond to states $t_n$ of random variables $g(t_1, \ldots, t_{n-1})$. When considering models over probabilistic logic programs one must guarantee that each random variable has at most one value, Therefore, Ngo and Haddawy need to introduce so called exclusivity constraints, such as $\leftarrow neighbourhood(X, average), neighbourhood(X, bad)$. These are unnecessary for Bayesian logic programs.

Secondly, Ngo and Haddawy employ an inference procedure that is exponentially slower than ours to construct the knowledge base. This is best illustrated on an example. Consider computing the density $p(burglary(\texttt{james}))$. Ngo and Haddawy's inference engine would construct one successful path for each possible value of all possible random variables influencing $burglary(james)$, as shown in Figure 13. In contrast, using Bayesian logic programs we would consider only one successful path. $burglary(\texttt{james}) \leftarrow neighbourhood(\texttt{james})$. For this simple example , this is a reduction by a factor 4. It is easy to show that, if we assume binary domains, Bayesian logic programs are exponentially more efficient.

Thirdly, in the probability models of Ngo and Haddaway, "*each random variable can assume a value from a finite set*" [NH97, page 149], i.e. that no (infinite) discrete or continuous random variables can be considered. Their inference procedure cannot cope with such variables. Furthermore, it is not entirely clear to what extent Ngo and Haddawy deal with function symbols[9]

Fourth, in probabilistic logic programs it is possible to employ partially defined associated densities (as in the example above), i.e. some entries in the conditional probability densities are undefined. Though Bayesian logic programs could - in principle - be extended to allow for such partially defined densities[10], we prefer not to do so. Reasons for this are that it complicates the notation and also that it is unclear whether there are any advantages of using such partially defined densities, cf. the ongoing discussion in the literature (e.g. [NH97,Jae98]).

Fifth, probabilistic logic programs are much more complex than Bayesian logic programs. They mix the qualitative information, the logical component with the quantitative information, whereas in Bayesian logic programs this information is - as in Bayesian networks - nicely separated. This separation of the two components is often considered one of the most important advantages of Bayesian networks.

A further extension of probabilistic logic programs is the use of context information. Context information is used to filter away sentences that do not apply to the current query from the knowledge base. Consider e.g. the following program (inspired by [NH97])

$$P(neighbourhood(X, bad)) = 0.2 \leftarrow lives\_in(X, yorkshire)$$
$$P(neighbourhood(X, bad)) = 0.4 \leftarrow lives\_in(X, vienna)$$

It states that we have different conditional probabilities depending on whether $lives\_in(x, yorkshire)$ or $lives\_in(x, vienna)$ is true or false, given some external logic program[11]. Whereas context information may be important for efficiency reasons, we believe it is more natural to view this information as deterministic knowledge that can be specified using a "pure" Prolog program (a subset of Bayesian logic programs) although the filter process used in [NH97] can easily be incorporated into our framework. Using this approach, the previous probabilistic logic programs would be written as the following Bayesian logic program

```
neighbourhood(X) | lives_in(X,yorkshire)
neighbourhood(X) | lives_in(X,vienna)
```

where we assume the existence of some "pure" Prolog clauses defining `lives_in`.

---

[9] [NH97] contains a theorem without proof about countable infinite "least Herbrand domains" (i.e. RAS). However, in their main paper [NH97], Ngo and Haddawy define the semantics only for the case that the "least Herbrand domain" is finite.

[10] Alternatively, we could handle partially defined conditional probability densities using the combining rules.

[11] The external logic programming may employ negation, i.e. it is a so called normal logic program. We cannot deal with negation as failure or completion semantics but we could use Bayesian logic programs to define negated predicates explicitly.

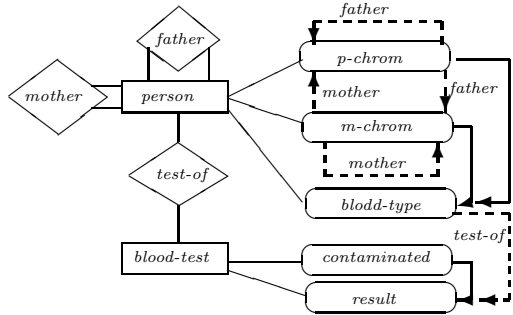

Figure 14. Probabilistic relational model of a genetic domain. We use the standard graphical notation of entity/relationship models: ovals represent attributes and boxes entities. Dashed lines indicate aggregations as parents, solid ones indicate attributes as parents.

To summarize, Bayesian and probabilistic logic programs are strongly related. However, Bayesian logic programs are simpler, more natural, more efficient and more expressive. They are simpler as they are based on less concepts and notation; they are more natural as they clearly separate the logical from the probabilistic component; and they are more expressive as they can represent functors and continuous variables.

Probabilistic and Bayesian logic programs are also related to Poole's framework of probabilistic Horn abduction [Poo93], which is "*a pragmatically-motivated simple logic formulation that includes definite clauses and probabilities over hypotheses*" [Poo93]. Poole's framework provides a link to abduction and assumption-based reasoning. However, as Ngo and Haddawy point out, probabilistic and therefore also Bayesian logic programs have not as many constraints on the representation language, represent probabilistic dependencies directly rather than indirectly, have a richer representational power, and their independency assumption reflects the causality of the domain.

## 8.2. Probabilistic relational models

Koller et. al. [FGKP99,Kol99] define probabilistic relational models, which are based on the well-known entity/relationship model. Figure 14 shows an example from [FGKP99]: "it is a genetic model of the inheritance of a single gene that determines a person's blood type. Each person has two copies of the chromosome containing this gene, one inherited from her mother, and one inherited from her father. There is also a possibly contaminated test that attempts to recognize the person's blood type." In probabilistic relational models, the random variables are the attributes. The relations between entities are deterministic, i.e. they are only true or false. To represent this within Bayesian logic programs Koller et al. use the following normal form: each attribute $a$ of an entity type $E$ is a Bayesian predicate $a(E)$ and each $n$-ary relation $r$ is a $n$-ary logical Bayesian predicate $n$. Probabilistic relational models consist of a qualitative dependency structure over the attributes and their associated quantitative parameters (the conditional probability densities). Koller et. al. distinguish among two types of parents of an attribute. First, an attribute $a(X)$ can depend on another attribute $b(X)$, e.g. the blood type of a person depends on the chromosome inherited from its father $(p - chrom)$. This is equivalent to the Bayesian clause `a(X) | b(X)`. Second, an attribute $a(X)$ possibly depends on an attribute $b(Y)$ of an entity $Y$ related to $X$, e.g. the chromosomes of a person depends on the chromosomes of its mother $(m - chrom)$. The relation between $X$ and $Y$ is described by a slot $s(X, Y)$ which is either a projection of a relation, i.e. `s(X,Y) :- r(X₁,...,Xₙ)`, or a composition of slots, i.e. `s(X,Y) :- s₁(X,X₁),s₂(X₁,X₂),...,sₘ(X_{m-1},Y)`. Given these logical Bayesian clauses, the original dependency is represented by `a(X) | s(X,Y),b(Y)`. Thus, the example in Figure 14 can be represented as the Bayesian logic program

```
m_chrom(X)     | mother(X,Y), p_chrom(Y), m_chrom(Y).
p_chrom(X)     | father(X,Y), p_chrom(Y), m_chrom(Y).
blood_type(X)  | m_chrom(X), p_chrom(X).
contaminated(X)| blood_test(X).
result(X)      | test_of(X,Y), contaminated(X), blood_type(Y).
```

One original feature of probabilistic relational models concerns the way they deal with multiple instantiations of a single clause. To this purpose Bayesian logic programs employ combining rules. Probabilistic relational models however use aggregate functions (as in database languages) to map multiple values of an attribute onto a single value. E.g. consider the above clause for `m_chrom` and assume that there are multiple possible values for $Y$. Then the probabilistic relational model would apply an aggregate function to these values and specify a probability density conditioned over the derived attributes. If one would desire to use aggregate functions within Bayesian logic programs one has to simulate this calculation by moving the aggregate function inside the combining rules. The input of a combining rule is a set of conditional probability densities specifying partial influences. Indeed, the associated conditional probability densities of a Bayesian logic program list the involved random variables as well as their values, so that the combining rule could implicitly employ the aggregate functions to realize the same effects. It has to compute the aggregation for all possible states of involved random variables.

At this point it should be clear that probabilistic relational models employ a more restricted logical component than Bayesian logic programs do: The component is a restricted version of the commonly used entity/relationship model, where relations have attributes, and any entity/relationship model can be represented using a (range-restricted) definite clause logic. Finally, let us point at one difficulty is introduced by the concept of slot chains, as slots are by definition binary relations. It is well-known from database theory that − in general − a ternary relation cannot be represented using binary relations only, (see e.g. [EN94]). Suppose, for instance that we have the ternary relation *beer_drinker*[12]

| | *beer_drinker* | |
|---|---|---|
| *pub* | *guest* | *beer* |
| Kowalski | Kemper | Pils |
| Kowalski | Eickler | Hefeweizen |
| Innsteg | Kemper | Hefeweizen |

The slot chain *beer_drinker*(*pub*, *guest*) ∘ *beer_drinker*(*guest*, *beer*) yields the relation

| | *beer_drinker*(*pub*, *guest*) ∘ *beer_drinker*(*guest*, *beer*) | |
|---|---|---|
| *pub* | *guest* | *beer* |
| Kowalski | Kemper | Pils |
| Kowalski | Kemper | Hefeweizen |
| Kowalski | Eickler | Hefeweizen |
| Innsteg | Kemper | Pils |
| Innsteg | Kemper | Hefeweizen |

where spurious tuples bearing wrong information are introduced. This will also be the case for any other projection of the ternary relation. As a consequence, it is unclear how to handle such ternary relations using probabilistic relational models.

## 8.3. Relational Bayesian networks

Jaeger [Jae97] considers Bayesian networks where the nodes are predicate symbols. The states of these random variables are possible interpretations of the symbols over an arbitrary, finite domain (here we only consider Herbrand domains), i.e. the random variables are set-valued. On the other hand, the inference problem addressed by Jaeger does not ask for the probability of a ground atom belonging to any specific interpretation, but only for the probability that an interpretation contains that ground atom. Under these conditions, relational Bayesian networks are viewed as Bayesian networks where the nodes are the ground atoms and all random variables have the domain {*true*, *false*}[13]. The key difference between relational Bayesian networks and Bayesian logic programs is that the quantitative information is specified by so called probability formulas. These formulas employ the notion of combination functions, functions that map every finite multiset with elements from [0, 1] into [0, 1], as well as that of equality

---

[12] The example is taken from [KE97, page 153].
[13] It is possible, but complicated to model domains having more than two values.

constraints[14]. Let $F_{cancer}(x)$ be

$$noisy\text{-}or\{comb_\Gamma\{exposed(x, y, z) \mid z; true\} \mid y; true\}$$

This formula states that that for any specific organ $y$, multiple exposures to radiation have a cumulative effect on the risk of developing cancer of $y$. But developing cancer at any of the various organs $y$ can be viewed as independent causes. As shown in [Jae97] a probability formula not only specifies the densities but also the dependency structure. Because of this and the computational power of combining rules, a probability formula is easily expressed as a set of Bayesian clauses: the head of the Bayesian clauses is the corresponding Bayesian atom and the bodies consist of all maximally generalized Bayesian atoms occurring in the probability formula. Now the combining rule can select the right ground atoms and simulate the probability formula on them. This is always possible because the Herbrand base is finite. E.g. the clause `cancer(X) | exposed(X,Y,Z)` together with the right combining rule and associated conditional probability densities models the example formula. We refer for a more detailed discussion to [Ker00].

## 9. Conclusion

We have introduced Bayesian logic programs, their representation language, their declarative semantics and their query-answering procedure. Bayesian logic programs are a novel framework for combining Bayesian networks with definite clause logic. The main idea of Bayesian logic programs is to establish a one-to-one mapping between true (logical) ground atoms and random variables. This ensures a unique probability density over the random variables corresponding to the least Herbrand model. The least Herbrand model of a Bayesian logic program together with its direct influence relation is interpretable as a (possibly infinite) Bayesian network where the parents of a random variable $X$ are (Bayesian) ground atoms directly influencing $X$. The query-answering procedure adapts the two phase strategy of *knowledge-based model construction* methods: (1) construct the support network, and (2) apply a Bayesian network (exact or approximating) inference algorithm on the support network. The procedure is based on AND/OR graphs which allow to intuitively combine the logical and the probabilistic information. It turned out that even if the least Herbrand model is infinite every well-defined probabilistic query is computable.

We also argued that Bayesian logic programs inherit the advantages of both Bayesian networks and definite clause logic, including the strict separation of qualitative and quantitative aspects. Indeed, Bayesian logic programs can naturally model any type of Bayesian network (including those involving continuous variables) as well as any type of "pure" Prolog program (including those involving functors). Therefore Bayesian logic programs should be easy to use and understand by people familiar with the underlying representations. We also demonstrated that Bayesian logic programs can model dynamic Bayesian networks and hidden Markov models and investigated their relationship to other first order extensions of Bayesian networks. Finally, a simple Prolog meta-interpreter was presented that performs knowledge base construction for Bayesian logic programs.

---

[14] To simplify the discussion, we will further ignore these equality constraints here. For details we refer to [Ker00].

In the future, we plan to apply Bayesian logic programs to some real-world problems. Also, the learning of Bayesian logic programs will be investigated. We believe that the strict separation property will also be advantageous in this respect as inductive logic programming techniques [MD94] could be applied to learn the structure of the Bayesian logic program and probabilistic Bayesian network techniques could be applied to determine the parameters of the network. Some preliminary suggestions concerning learning can be found in [KDK00]. Further work may also be concerned with more efficient inference algorithms, along the lines of e.g structured inference [Pfe99] or [KMP97]. Finally, the relation between Bayesian logic programs and other probabilistic first order logics, such as Muggleton's stochastic logic programs [Mug96], could be investigated. This question seems to be related to the difference between Halpern's type I and type II probabilistic structures [Hal89].

## 10. Acknowledgements

We would like to thank James Cussens, Peter Flach, Manfred Jaeger, Daphne Koller, Stefan Kramer, Stephen Muggleton and David Page for discussions and encouragement on Bayesian logic programs. Many of the graphs in this paper were drawn using the program *dot* [KN99].

# Appendix

## A. The Mathematical Background of the Proofs of Theorem 4.9 and Theorem 5.3

Here we introduce the concepts of probability theory needed to prove Theorem 4.9 and Theorem 5.3. More information can be found e.g. in [Bau91,Bau92, FG97].

A system $\mathcal{A}$ of subsets of a set $\Omega$ is a *$\sigma$-algebra* (in $\Omega$) if it exhibits the following features: (1) $\Omega \in \mathcal{A}$; (2) if $A \in \mathcal{A} \rightarrow \bar{A} \in \mathcal{A}$, where $\bar{A}$ denotes the complementary set of $A$; (3) for each sequence $(A_n)$, $A_n \in \mathcal{A}$, the proposition $\bigcup_{n=1}^{\infty} A_N \in \mathcal{A}$ holds. E.g. the power set of a given set is a $\sigma$-algebra. A *probability space* is a tuple $(\Omega, \mathcal{A}, P)$ where $\Omega$ is a set, $\mathcal{A}$ is a $\sigma$-algebra in $\Omega$ and $P : \Omega \mapsto [0, 1]$ a *probability measure*. A *probability measure* $P$ (over $(\Omega, \mathcal{A})$) satisfies (1) $P(\emptyset) = 0$, (2) for each sequence $(A_n)$ of pairwise disjunctive sets in $\mathcal{A}$, whose union lies in $\mathcal{A}$, the equation $P(\bigcup_{n+1}^{\infty} A_n) = \sum_{n=1}^{\infty} P(A_n)$ holds and (3) $P(\Omega) = 1$. One can show that for all $A \in \mathcal{A} : 0 \leq P(A) \leq 1$. The elements of $\Omega$ are called the *elemental* events, and the elements of $\mathcal{A}$ are called *events*; the empty set is the *impossible* and $\Omega$ the *certain* event. The value $P(A)$ for $A \in \mathcal{A}$ is called the *probability* of $A$. A pair of a set and a $\sigma$-algebra over that set is called a *measurable space*, e.g. $(\Omega, \mathcal{A})$ is a measurable space. A mapping $T : \Omega \rightarrow \Omega'$ where $(\Omega, \mathcal{A})$ and $(\Omega', \mathcal{A}')$ are measurable spaces, is called $(\mathcal{A} - \mathcal{A}')$-*measurable* if for all $A' \in \mathcal{A}$: $T^{-1}(A') \in \mathcal{A}$. Let $(\Omega', \mathcal{A}')$ be a measurable space. A *random variable* $X$ with domain $\mathrm{D}(X) = \Omega'$ is a $(\mathcal{A} - \mathcal{A}')$-measurable function $X : \Omega \mapsto \Omega'$. The *distribution* of $X$ is the image measure $X(P)$. In the case of $(\Omega', \mathcal{A}') = (\mathbb{R}, \mathcal{B}^1)$ we call $X$ a *real* random variable where $\mathcal{B}^1$ is the minimal $\sigma$-algebra which contains all half-open intervals $[a, b) \subset \mathbb{R}$. Let $I$ be a countable index set, $(x_n)_{n \in I}$ a sequence in $\mathbb{R}$ and $(\alpha_n)_{n \in \mathbb{N}}$ a sequence of non-negative real numbers with $\sum_{n=1}^{\infty} \alpha_n = 1$. A *discrete* random variable is a real random variable having $P = \sum_{n=1}^{\infty} \alpha_n \epsilon_{x_n}$ as distribution where $\epsilon_{x_n}(A) = 1$ if $x_n \in A$ and $\epsilon_{x_n}(A) = 0$ if $x_n \notin A$. In the case of $I$ being finite $X$ would be called a *finite* random variable.

Let $T$ be a countable index set. A real *discrete-time stochastic process* $S$ is a tuple $S = (\Omega, \mathcal{A}, P, (X_t)_{t \in T})$ where $(\Omega, \mathcal{A}, P)$ is a probability space and $(X_t)_{t \in T}$ is a family of real random variables over this probability space . Let $J \subset T$ be finite. The distribution of $(X_t)_{t \in J}$, denoted as $P_J$ is defined to be $X_J(P)$ where $X_J$ is the product mapping $X_J := \bigotimes_{i \in J} X_t$ (cf. see [Bau91]). Let $\mathcal{H}(T)$ be the set of all non-empty, finite subsets of $T$. The family $(P_J)_{J \in \mathcal{H}(T)}$ is called the *family of finite-dimensional distributions* of $S$. Kolmogorov's theorem reads with respect to real discrete-time stochastic processes as follows:

**Theorem A.1** *(Kolmogorov). Let $T$ be a non-empty, countable set and $(\mathbb{R}^J, \mathcal{R}^J, P_J)_{J \in \mathcal{H}(T)}$ a family of probability spaces. If the family is* projective, *then there exists a unique probability measure $P_T$ over $(\mathbb{R}^T, \mathcal{B}^T)$ with $proj_J^T(P_I) = P_J$ for all $J \in \mathcal{H}(T)$. The family $(\mathbb{R}^J, \mathcal{R}^J, P_J)_{J \in \mathcal{H}(T)}$ is called* projective *if for all $H, J \in \mathcal{H}(T)$ with $H \subset J$ the equation $P_J = proj_J^H(P_H)$ holds, where $proj_J^H$ is the projection of $\mathbb{R}^H$ onto $\mathbb{R}^J$.*

If we interpret the index set $T$ as time steps, then "*the preceding result [...] says that the distribution of $(X_t)_{t \in T}$ can be specified by giving the 'initial distribution' (the distribution of $X_0$) and, for each time $n$, giving the conditional distribution of the 'present' value $X_n$ in term of the 'past' values $X_0, \ldots, X_{n-1}$*" [FG97, page 433].

In the proofs of Theorem 4.9 and of Theorem 5.3 we will use one of the key features of Bayesian networks. An induction over the size of a Bayesian network $N$ over the real random variables $X_1, \ldots, X_l$, $l > 0$ shows that $N$ specifies a projective family of probability measures. Thus, it holds

$$\mathbf{p}(\mathbf{J}) = \int_{-\infty}^{+\infty} \ldots \int_{-\infty}^{+\infty} \mathbf{p}(\mathbf{H}) \, d\mathbf{D}$$

where $\mathbf{J} \subset \mathbf{H} \subset N$ and $\mathbf{D} = \mathbf{H} \setminus \mathbf{J}$.

## B. Interpreting Bayesian Logic Programs using Prolog - BLoP

Here we present a rudimentary interpreter of Bayesian logic programs in Prolog. It is intended to get a "playfully" understanding of Bayesian logic programs. We would like to point out that the interpreter is by no means optimal. It is just a simple and direct implementation of the query-answering procedure. Therefore, we have neither implemented pruning nor optimization mechanisms in constructing the SLD tree.

The interpreter resembles the first phase of the query-answering procedure in Section 6.2: It builds the support network with respect to a given Bayesian logic program and a well-defined probabilistic query. The support network is specified in the Hugin net file language (see [Hugb,Hug01]) and written to the file 'out.net'. It can directly be loaded into the HUGIN system (a demo version of Hugin can be downloaded from `www.hugin.com`), although the file `out.net` itself is well readable for humans. The heart of the interpreter (`prove/3`) is a common Prolog meta interpreter as one can find in every standard introductory text book on Prolog (e.g. [SS86,Bra86,Fla94]). The code was developed under Sicstus 3.8.1. It is best explained using an example.

Assume, we want to answer the query `?- height(fred) | height(ann)=159` to the Bayesian logic program *height*, i.e. we need the support network $N(height(henry), height(irene))$. The *height* program can be formulated as follows:

```
1   domain(father/2,discrete,[true, false]).
2   domain(mother/2,discrete,[true, false]).
3   domain(height/1,continuous,real).
```

We have two Bayesian predicates `father/2` and `mother/2` having both the (finitely) discrete domain {*true, false*}. Furthermore, `height/1` is a Bayesian predicate with a continuous domain, the real numbers. We use the usual Prolog notation `p/N` to refer to a $N$-ary predicate $p$.

```
4   combining_rule(father/2,id).
5   combining_rule(mother/2,id).
6   combining_rule(height/1,id).
```

The associated combining rules are the identity. Other combining rules should be easily defined (see below). The qualitative aspects of the *height* domain are specified as follows:

```
7    father(unknown1,fred).  mother(ann, fred).   father(brian,dorothy).
8    mother(ann, dorothy).   father(brian,eric).  mother(cecily,eric).
9    father(unknown2,gwenn). mother(ann,gwenn).   father(fred,henry).
10   mother(dorothy,henry).  father(eric,irene).  mother(gwenn,irene).
11   father(henry,john).     mother(irene,john).
12
13   height(ann).            height(brian).  height(cecily).
14   height(unknown1).       height(unknown2).
15
16   height(X) | mother(Y,X),father(Z,X),height(Y),height(Z).
```

The associated conditional probability densities are specified using `cpd/2`. The first argument is the corresponding Bayesian clause. The second argument are the densities. (Finitely) discrete densities are represented as a list of prolog terms, whereas conditional Gaussian densities are represented by a list of ground atoms of the form `normal(m,v)`. The terms `m` and `v` refer to the mean and the variance of a Gaussian density. The notation is based upon the HUGIN net specification language and we refer for a discussion to [Hugb,Hug01]. We believe it should be easy to extend the supported densities with respect to the HUGIN net language.

```
17   cpd(father(unknown1,fred),[1.0,0.0]).      cpd(mother(ann, fred),[1.0,0.0]).
18   cpd(father(brian,dorothy),[1.0,0.0]).      cpd(mother(ann, dorothy),[1.0,0.0]).
19   cpd(father(brian,eric),[1.0,0.0]).         cpd(mother(cecily,eric),[1.0,0.0]).
20   cpd(father(unknown2,gwenn),[1.0,0.0]).     cpd(mother(ann,gwenn),[1.0,0.0]).
21   cpd(father(fred,henry),[1.0,0.0]).         cpd(mother(dorothy,henry),[1.0,0.0]).
22   cpd(father(eric,irene),[1.0,0.0]).         cpd(mother(gwenn,irene),[1.0,0.0]).
23   cpd(father(henry,john),[1.0,0.0]).         cpd(mother(irene,john),[1.0,0.0]).
24
25   cpd(height(ann),[normal(165,60)]).         cpd(height(brian),[normal(165,60)]).
26   cpd(height(cecily),[normal(165,60)]).      cpd(height(unknown1),[normal(165,60)]).
27   cpd(height(unknown2),[normal(165,60)]).
28
29   cpd((height(X)|mother(Y,X),father(Z,X),height(Y),height(Z)),
30      [normal(0.5*height(Y)+0.5*height(Z),0),
31       normal(165,60),normal(165,60),normal(165,120)]).
```

We will now describe the core of the BLoP interpreter. The BLoP interpreter could be started using `blop_shell`. A little shell for computing support networks is opened. First, we have to consult the *height* Bayesian logic program, [`'height.blp'`]. Then, the support network $N(height(fred), height(ann))$ is computed by typing `height(fred) | height(ann)=155.` at the shell prompt. The algorithmic skeleton (see algorithm 6.2) is realized by `query/2`.

```
1    query(Q,E) :- !,
2       retractall(proof(_)), retractall(network(_)),
3       assert(proof([])),
4       evidence_variables(E,EVars), append(Q,EVars,Vars),
5       solution_graph(Vars,S),
6       support_network(S,SNC),
7       apply_comb(SNC,SN), asserta(network(SN)),
8       write_support_network(SN).
9
10   evidence_variables([],[]).
11   evidence_variables([V=_|Vs],[V|Vars]) :-
12      evidence_variables(Vs,Vars).
```

After extracting the random variables occurring in the evidence, `evidence_variables(E,EVars)`, `query/2` computes the solution graph `S` for the variables `Vars` calling `solution_graph(Vars,S)`.

```
13   prove(true,P,P):- !.
14   prove((GA,GB),OP,NP) :- !,
15      prove(GA,OP,PA), prove(GB,PA,NP).
16   prove(G,OP,[G-B|P]) :-
17      clause(G,B), prove(B,OP,P).
```

```
18
19  solution_graph([G|Gs],S) :-
20    prove(G,[],P), assertz(proof(P)), fail;
21    solution_graph(Gs,S);
22    assertz(proof(end)), collect(US), sort(US,S).
23
24  collect(L) :-
25    retract(proof(X)),!,
26    (X == end,!,L=[];
27      collect(R), append(X,R,L).
```

The Prolog procedure `solution_graph/2` works similar to `findall/3` as defined in [Bra86]. The output `S` is a lexicographically ordered Prolog list of edges `A-B` from an *or* node `A` to an *and* node `B` in the solution graph of `Vars`. To compute the successful paths we have adapted a depth first searching meta interpreter `prove/3` as described in [SS86]. It returns the list of successful paths in the last argument. Having the solution graph `S`, the Prolog procedure `support_network/2` looks up the associated conditional probability densities of each clause used to build `S`:

```
28  support_network([],[]).
29  support_network([N-Pa|Es],[(N,T,D,[Parents|Pas],[CPD|CPDs])|SN]) :-
30    functor(N,P,NA), domain(P/NA,T,D),
31    cpd(N,Pa,Parents,CPD),
32    support_network(N,Es,Pas,CPDs,Rest),
33    support_network(Rest,SN).
34  support_network(_,[],[],[],[]).
35  support_network(N,[N-Pa|R],[Parents|Pas],[CPD|CPDs],Rest) :-
36    cpd(N,Pa,Parents,CPD),
37    support_network(N,R,Pas,CPDs,Rest).
38  support_network(_,R,[],[],R).
```

The last step is to compute the combined conditional probability densities. This is done using `apply_comb/2`:

```
39  apply_comb([],[]).
40  apply_comb([F|Fs],[CF|CFs]) :-
41    combine(F,CF),
42    apply_comb(Fs,CFs).
43
44  combine((N,R,T,D,LPa,LCPDs),(N,R,T,D,Pa,CPD)) :-
45    functor(N,P,NA),
46    clause(combining_rule(P/NA,CR),_),
47    C=..[CR,LPa,LCPDs,Pa,CPD],
48    call(C).
```

The corresponding combining rule of the Bayesian predicate `P` is called by `call(C)`. A combining rule `cr` is seen as a Prolog procedure `cr(PaL,CPDL,Pa,CPD)`. The `PaL` argument is the given list of set of parents. The `CPDL` argument specifies the corresponding conditional probability densities. Thus, the $i$-th element of `CPDL` are the conditional probability densities associated to the $i$-th parent set in `PaL`. The resulting set of parents and the combined conditional probability densities are returned in `Pa` and `CPD`.

After performing all steps, the support network $N(height(fred), height(ann))$ is given in the variable SN. It will be written out to the file out.net. The computed support network $N(height(fred), height(ann))$ equals the support network $N(height(fred))$ shown Figure 7). The rest of the code implements the shell and input/output operation. Because the way it works gives no insights in the interpreter, we will not explain it.

```
49  %
50  :- use_module(library(lists)).
51  :- use_module(library(charsio)).
52  :- op(500,xfy,'|').
53  :- dynamic domain/3, combining_rule/2, network/1.
54
55  %
56  term_expansion((Head|Body1,Body2), (Head:-Body1,Body2)).
57  term_expansion((Head|Body), (Head:-Body)).
58  term_expansion(domain(A/N,T,D),(:- dynamic A/N)) :-
59          retractall(domain(A/N,_,_)),assert(domain(A/N,T,D)).
60  term_expansion(combining_rule(A/N,B),(:- dynamic A/N)) :-
61          retractall(combining_rule(A/N,_)),assert(combining_rule(A/N,B)).
62  term_expansion(cpd((H|B),CPD),cpd(H,B,BL,CPD)) :-
63          conj_2_list(B,BL).
64  term_expansion(cpd(H,CPD),cpd(H,true,[],CPD)) :- !.
65
66  conj_2_list((A,B),L) :-
67          conj_2_list(A,LA),
68          conj_2_list(B,LB),
69          append(LA,LB,L).
70  conj_2_list((A),[A]).
71
72  %
73  blop_shell :-
74      assert(network([])), blop_shell_help, blop_shell(next).
75  blop_shell(next) :- !,
76      blop_shell_prompt, char_conversion('|', ';'),
77      read(Goal), char_conversion('|', '|'), blop_shell(Goal).
78  blop_shell(exit).
79  blop_shell_help :- !,nl,
80      write('   ********************************************************'),nl,
81      write('   *        BLoP - Bayesian Logic Programs Interpreter       *'),nl,
82      write('   ********************************************************'),nl,
83      write('   * help.                    |+ this message              *'),nl,
84      write('   * Q1,...QN.                |+ support network for computing *'),nl,
85      write('   *                          |  p(Q1,...,QN)               *'),nl,
86      write('   * Q1,...,QN|E1=e1,...,EM=eM. |+ support network for computing *'),nl,
87      write('   *                          |  p(Q1,...,QN|E1=e1,...,EM=eM) *'),nl,
88      write('   * [''height.blp''].          |+ consult BLP file ''height.blp'' *'),nl,
89      write('   * exit.                    |+ exit the shell            *'),nl,
90      write('   ********************************************************'),nl,nl.
91  blop_shell(help) :-
92      blop_shell_help, blop_shell(next).
93  blop_shell([X]) :-
94      consult(X),!, blop_shell(next).
95  blop_shell((Q;E)) :-!,
96      query([Q],[E]), blop_shell(next).
97  blop_shell(Q) :-
98      query([Q],[]),!, blop_shell(next).
```

```
99   blop_shell(_) :- !,
100      format('~nUnknown command !~n~n',[]), blop_shell(next).
101  blop_shell_prompt :-
102      write('<BLoP> ?- '), flush_output(user_output).
103
104  %
105  id([Pa],[CPD],Pa,CPD).
106
107  %
108  write_support_network(SN):-
109      flush_output(user_output), tell('out.net'),
110      format('net~n{~n  node_size = (100 40);~n}~n~n',[]),
111      write_nodes(SN), write_potentials(SN), told.
112
113  write_nodes([]).
114  write_nodes([[N,T,D,_,_]|R]) :- write_node(N,T,D), write_nodes(R).
115
116  write_node(N,continuous,_) :-
117      format('continuous node ~@~n{~n  label = "~q";~n}~n~n',[write_no_brackets(N),N]).
118  write_node(N,discrete,D) :-
119      format('discrete node ~@~n{~n  states = ( ~@ );~n  label = "~q";~n}~n~n',
120          [write_no_brackets(N),write_ddomain(D),N]).
121
122  write_ddomain([]).
123  write_ddomain([D|Ds]) :- format('"~q" ',[D]), write_ddomain(Ds).
124
125  write_potentials([]).
126  write_potentials([[N,continuous,_,Pa,CPD]|R]) :-
127      format('potential (~@ ~@)~n{~n  data = ( ~@ );~n}~n~n',
128          [write_no_brackets(N), write_cond(Pa), write_continuous_cpd(CPD)]),
129      write_potentials(R).
130  write_potentials([[N,discrete,_,Pa,CPD]|R]) :-
131      format('potential (~@ ~@)~n{~n  data = ( ~@ );~n}~n~n',
132          [write_no_brackets(N), write_cond(Pa), write_list(CPD)]),
133      write_potentials(R).
134
135  write_cond([]).
136  write_cond(P) :- format(' | ~@ ',[write_list(P)]).
137
138  write_continuous_cpd([]).
139  write_continuous_cpd([normal(M,V)|Ps]) :-
140      format(' normal(~@,~@) ',[write_no_brackets(M), write_no_brackets(V) ]),
141          write_continuous_cpd(Ps).
142
143  write_list([]).
144  write_list([P|Ps]) :-
145      format(' ~@ ',[        write_no_brackets(P) ]),write_list(Ps).
146
147  write_no_brackets(T) :-
148      write_to_chars(T,C), delete(C,40,C1), delete(C1,41,C2),
149      substitute(44,C2,95,NT), name(N,NT), write(N).
```